%% file: main.tex
\documentclass{applemlr}
\usepackage{amsmath}
\usepackage{enumerate} 
\usepackage{algorithm}
\usepackage{algpseudocode}
\usepackage{amsfonts}
\usepackage{amsthm}
\usepackage{newtxtt}
\usepackage{cleveref}
\usepackage{diagbox} 
\usepackage{colortbl}
\usepackage{amssymb}
\usepackage{xspace}
\usepackage{wrapfig}
\usepackage{adjustbox}
\usepackage{tabularx}
\usepackage{booktabs}
\usepackage{mathtools}
\usepackage{tikz}
\usepackage{enumitem}
\usepackage{silence}
\usepackage{dsfont}
\usepackage[table]{xcolor}
\usepackage[dvipsnames]{xcolor}
\usepackage{multirow}
\usepackage{makecell}
\usepackage{xfakebold}
\usepackage{pifont} 
\input{math_commands}

\usepackage{graphicx}

\usepackage{color}
\usepackage[dvipsnames]{xcolor}
\definecolor{JalapenoRed}{RGB}{183,21,64}
\definecolor{Belize}{RGB}{41,128,185}
\definecolor{Amour}{RGB}{238,82,83}
\definecolor{lightblue}{HTML}{dfebf7}
\definecolor{Gray}{gray}{0.93}
\definecolor{magnolia}{rgb}{0.97, 0.96, 1.0}
\definecolor{mayablue}{rgb}{0.45, 0.76, 0.98}
\definecolor{lavenderblue}{rgb}{0.8, 0.8, 1.0}
\definecolor{lavender}{rgb}{0.9, 0.9, 0.98}
\definecolor{islamicgreen}{rgb}{0.0, 0.56, 0.0}

\usepackage{stfloats}

\newcommand{\cmark}{\ding{51}}
\newcommand{\xmark}{\ding{55}}

\definecolor{textgray}{HTML}{6E6E73}
\usetikzlibrary{positioning, calc}
\usetikzlibrary{decorations.pathmorphing}

\makeatletter
\patchcmd{\wrong@fontshape}{\@gobbletwo}{}{}{}
\makeatother
\makeatletter
\AtBeginDocument{%
  \urlstyle{sf}
  
}
\makeatother

\numberwithin{equation}{section} 
\tcbuselibrary{minted}
\setminted[python]{
  linenos,
  breaklines,
  fontsize=\footnotesize,
  xleftmargin=2em
}

\definecolor{light}{RGB}{125, 125, 125}
\crefname{tcb@cnt@pbox}{code}{code}
\Crefname{tcb@cnt@pbox}{Code}{Code}
\crefname{assumption}{assumption}{assumption}
\Crefname{assumption}{Assumption}{Assumptions}

\newtcolorbox[auto counter]{pbox}[2][]{
  colback=white,
  title=Code~\thetcbcounter: #2,
  #1,fonttitle=\sffamily,
  fontupper=\sffamily,
  arc=2pt,
  colframe=bgcolor,
  coltitle=fgcolor,
  colbacktitle=bgcolor,
  toptitle=0.25cm,
  bottomtitle=0.125cm
}

\tcbuselibrary{breakable,skins}
\newtcolorbox[auto counter]{promptbox}[2][]{
  breakable,
  enhanced,
  colback=lavender,
  colframe=Belize,
  boxrule=0.5pt,
  arc=2pt,
  fonttitle=\sffamily\bfseries\small,
  coltitle=black,
  colbacktitle=lightblue,
  title=Prompt~\thetcbcounter: #2,
  #1,
  left=1.5mm,right=1.5mm,top=1mm,bottom=1mm,
}
\crefname{tcb@cnt@promptbox}{prompt}{prompts}
\Crefname{tcb@cnt@promptbox}{Prompt}{Prompts}

\makeatletter
\newcommand\applefootnote[1]{%
  \begingroup
  \renewcommand\thefootnote{}%
  \renewcommand\@makefntext[1]{\noindent##1}%
  \footnote{#1}%
  \addtocounter{footnote}{-1}%
  \endgroup
}
\makeatother

\definecolor{cverbbg}{gray}{0.90}

\newcommand{\benchname}{MM-ToolSandBox}

\title{{MM-ToolSandBox}: A Unified Framework for Evaluating Visual Tool-Calling Agents}

\author{
\parbox{\textwidth}{
Kaixin Ma$^*$, Di Feng$^*$, Alexander Metz, Jiarui Lu, Eshan Verma, Afshin Dehghan
}}

\affiliation{Apple}
\contribution{$^*$Equal Contribution}

\abstract{
We introduce \textbf{MM-ToolSandBox}, a benchmark and evaluation framework for visually grounded tool-calling agents. The framework provides a stateful execution environment spanning 500+ tools across 16 application domains, supporting multi-image, multi-turn tasks where agents must ground progressively arriving visual inputs into executable tool calls while handling realistic conversational phenomena (goal revisions, error corrections, state mutations). An automated scenario generation pipeline produces diverse, visually grounded scenarios through information-flow-guided planning and multi-stage quality filtering, yielding $258$ human-verified nominal scenarios and $50$ variants targeting interactive UI applications. Evaluating $12$ state-of-the-art models, from $4$B open-weight to frontier proprietary systems, shows that current models still lack robust visual tool-calling capability: even the best model achieves below $50\%$ success rate. Our failure analysis further reveals that visual precision, not only planning, is a primary bottleneck for capable models: $53\%$ of failures stem from incorrect information extraction from images despite otherwise correct task workflows. A planning-to-precision crossover emerges with scale: smaller models fail at deciding \emph{what to do}, while larger models fail at perceiving \emph{what they see}, suggesting fundamentally different research directions for improving models at different capability levels. The framework and the benchmark are publicly available at \url{https://github.com/apple/ml-mmtoolsandbox}.
}

\date{\sffamily\today}

\begin{document}

\maketitle

\input{sec/1_intro}
\input{sec/2_related_works}
\input{sec/3_framework}
\input{sec/4_benchmark}
\input{sec/5_scenario_generation}
\input{sec/6_experiments}
\input{sec/7_conclusion}
\section{Acknowledgment}
We thank Andrew Szot for invaluable suggestions and discussions.

\clearpage
\newpage
\bibliographystyle{plainnat}
\bibliography{biblio}

\clearpage
\newpage
\appendix
\input{sec/X_suppl}

\end{document}

%% file: math_commands.tex
\usepackage{amsmath,amsfonts,bm}

\def\eqref#1{equation~\ref{#1}}

\def\1{\bm{1}}

\DeclareMathAlphabet{\mathsfit}{\encodingdefault}{\sfdefault}{m}{sl}
\SetMathAlphabet{\mathsfit}{bold}{\encodingdefault}{\sfdefault}{bx}{n}



%% file: sec/1_intro.tex
\section{Introduction}
\label{sec:introduction}
Evaluating LLM-based agents on tool-augmented tasks has become an active area of research, with benchmarks targeting function calling~\citep{li-etal-2023-api,patil2025the}, stateful environment interaction~\citep{lu2025toolsandbox,trivedi2024appworld}, and multi-turn conversational dynamics~\citep{froger2026gaia,barres2025tau2benchevaluatingconversationalagents,xiu2026astrabenchevaluatingtooluseagent}. However, most of these frameworks remain text-centric, and cannot evaluate many realistic assistant tasks with visual inputs, for example, a user may share a screenshot to diagnose an issue or provide a photo of an event poster to schedule an event. 

Solving visual tool-calling tasks requires more than visual recognition. The agent must extract task-relevant information from images, map visual evidence to the correct tool or code action, execute that action against a stateful environment, and continue the interaction when the user's intent evolves. This makes visual tool calling a distinct agentic capability: the model must not only understand what is shown, but also decide how to act on it through tools.

Among the few benchmarks that incorporate visual inputs, images are typically provided as a static prefix in the initial query~\citep{wang2024gta}, or the tasks are limited to single-round user-agent interactions~\citep{xie2024osworld,rawles2025androidworld}. Moreover, many existing benchmarks operate with relatively small tool spaces, often containing only $15$--$100$ tools~\citep{froger2026gaia,wang2024gta,kong2025mobileworldbenchmarkingautonomousmobile,lu2025toolsandbox}. As a result, there is no unified framework and benchmark for systematically studying the visual tool-calling capability of multimodal agents under dynamic, multi-turn, and large-tool-space settings.

\begin{figure*}[t]
\centering
\includegraphics[width=1\textwidth]{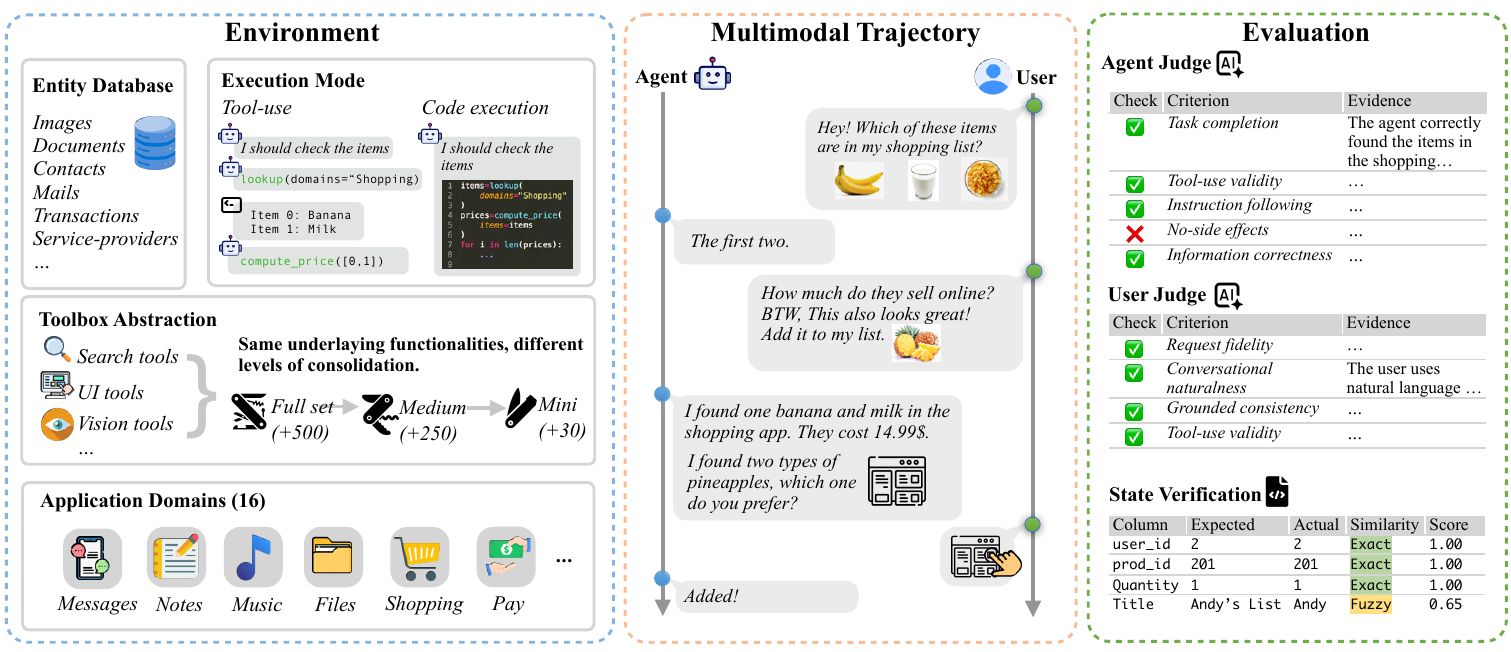}
\caption{Overview of the \benchname{} framework for evaluating visual tool-calling agents. The framework supports multi-turn, multi-image interaction between the user and agent, and provides both structured tool-use and code-execution interfaces. The framework includes over $500$ native tools across $16$ application domains, covering search, UI, vision, and system-level utilities. 
Evaluation combines rubric-based LLM judges for both agent and user roles with static state-verification checks.}
\label{fig:overview}
\end{figure*}

To close this gap, we introduce \textbf{\benchname{}}, a diverse simulation framework for evaluating visual tool-calling agents. \benchname{} extends the stateful tool-use environments from ToolSandBox~\citep{lu2025toolsandbox}, with explicit image handling, visual-specific tools, and support for both structured tool-use and code-execution interfaces, as summarized in Figure~\ref{fig:overview}. Built on top of this framework, we construct a visual tool-calling benchmark designed to capture the key challenges missing from prior work. The benchmark spans $511$ tools across $16$ simulated application domains and is organized along three orthogonal axes: seven information-flow types, four challenge types that model realistic conversational deviations, and four image-arrival patterns that control when visual inputs enter the interaction. The resulting benchmark contains $258$ nominal scenarios with $1{,}284$ unique images, requiring agents to perform multi-image, multi-turn tool use over a large and diverse tool space. All scenarios are further corrected and verified by human experts to ensure reliable visual grounding, executable task specifications, and well-defined completion criteria. We also curate $50$ scenarios to explore interactive UI applications. To make benchmark construction scalable while maintaining high quality, we develop an automated scenario generation pipeline. 
The pipeline produces visually grounded scenarios through information-flow-guided planning, environment-state instantiation, and multi-stage quality filtering, reducing the annotation burden from full manual construction to targeted human review. Finally, we conduct a systematic evaluation of state-of-the-art proprietary and open-weight models on our benchmark. Our results show that even strong frontier models struggle substantially with visual tool calling, revealing persistent gaps in image-grounded reasoning, multi-step interaction, and stateful tool use. In-depth analysis further shows that multi-image working memory is a key bottleneck, and that models of different scales exhibit distinct failure patterns. 

To summarize, our contributions are the following.
1. A stateful, diverse, multimodal agent simulation framework. We substantially extend ToolSandbox~\citep{lu2025toolsandbox}, a text-only environment with $34$ tools, into a multimodal runtime with persistent image artifacts that remain addressable across turns, dynamic image arrival during the conversation, $511$ tools across $16$ application domains with on-demand tool discovery, both code-execution and structured tool-use interfaces, and deterministic entity-diff verification for multimodal tasks.
2. An automated scenario-generation pipeline. The pipeline performs information-flow-guided planning, expands outlines into realistic multi-image, multi-turn interactions with executable environment states. In this way, we reduce benchmark construction from full manual authoring to targeted human review.
3. A human-verified benchmark and new empirical findings. Beyond leaderboard results, our analysis reveals a planning-to-precision failure crossover with model scale, in which smaller models fail predominantly at planning while stronger models increasingly fail at visual precision, and identifies multi-image working memory as a major bottleneck, with upfront image arrival roughly $20$ points harder than progressive arrival.

We release the benchmark and the framework to the public at \url{https://github.com/apple/ml-mmtoolsandbox}.

%% file: sec/2_related_works.tex
\section{Related Work}
\label{sec:related_work}
\noindent\textbf{Tool-use agent benchmarks.}
A growing body of benchmarks evaluate LLM agents on tool-augmented tasks. API-Bank~\citep{li-etal-2023-api} and BFCL~\citep{patil2025the} target single- and multi-turn function calling with large API collections. ToolSandbox~\citep{lu2025toolsandbox} introduces stateful environments with online evaluation and conversational simulation. AppWorld~\citep{trivedi2024appworld} scales to 457 APIs across 9 applications with rich relational data, though exclusively in a code-execution paradigm. More recent efforts push toward dynamic environments: Gaia2~\citep{froger2026gaia} introduces asynchronous, event-driven evaluation testing temporal awareness and agent-to-agent collaboration; $\tau^2$-Bench~\citep{barres2025tau2benchevaluatingconversationalagents} models agent-supervisor interactions revealing communication as a critical bottleneck; and ASTRA-bench~\citep{xiu2026astrabenchevaluatingtooluseagent} grounds evaluation in longitudinal personal context. GTA~\citep{wang2024gta} is among the first to incorporate multimodal contexts, but covers only 14 tools in a stateless setting. SO-Bench~\citep{feng2026sobenchstructuraloutputevaluation} evaluates the visual structural output capability of tool-use agents, but only focuses on single-turn interactions. Toolathlon~\citep{li2025toolathlon} further scales tool-use evaluation to long-horizon real-software workflows, spanning 32 software applications and 604 tools with realistic initial states and execution-based task verification. 
MCP-Atlas~\citep{bandi2026mcpatlas} evaluates agents on 1{,}000 multi-step tasks over 36 real MCP servers and 220 tools, requiring agents to discover and orchestrate 3--6 tool calls across servers from natural-language instructions. Despite this progress, these benchmarks operate predominantly in text-only regimes, evaluate under a single execution mode, and rely heavily on manual scenario construction.

\noindent\textbf{Multimodal tool-use agent benchmarks.}
A closer line of work evaluates agents that call tools over visual inputs. VisualToolBench~\citep{guo2025seeingevaluatingmultimodalllms}, TIR-Bench~\citep{li2025tirbenchcomprehensivebenchmarkagentic}, Agent-X~\citep{ashraf2026agentxevaluatingdeepmultimodal} and  AgentVista~\citep{su2026agentvistaevaluatingmultimodalagents} benchmark multimodal tool-using agents over a curated tool suite. These benchmarks establish that visual tool use is a distinct capability worth measuring, but they are predominantly single-turn and stateless, formulated as question answering, supply images as a static prefix to the task, and operate over a fairly small tool spaces. \benchname{} complements existing benchmarks with: a stateful environment in which tool calls mutate persistent world state, multi-turn interaction with a simulated user whose intent can change, dynamic image arrival across turns rather than a fixed visual prefix, a large tool space with over $500$ tools dual code-execution and structured tool-use interfaces, and evaluation by entity-level state verification rather than answer matching.



\noindent\textbf{Visual agent benchmarks.}
A parallel line evaluates agents that perceive and act on GUI inputs, known as GUI agents~\citep{wang2025ui,li2025ferret,yang2025ferret}. VisualWebArena~\citep{koh-etal-2024-visualwebarena} and WebVoyager~\citep{he-etal-2024-webvoyager} evaluate web navigation grounded in visual page understanding. At the OS level, OSWorld~\citep{xie2024osworld}, AndroidWorld~\citep{rawles2025androidworld} and UINavBench~\citep{agrawal2025uinavbench} benchmark desktop and mobile tasks, respectively, while MobileWorld~\citep{kong2025mobileworldbenchmarkingautonomousmobile} extends to long-horizon cross-application workflows. In multimodal search, MMSearch~\citep{jiang2024mmsearchbenchmarkingpotentiallarge} and MM-BrowseComp~\citep{li2025mmbrowsecompcomprehensivebenchmarkmultimodal} evaluate deep retrieval requiring visual-language reasoning. These benchmarks share grounding in visual perception but differ in image types (UI screenshots vs.\ natural images), action spaces (GUI operations vs.\ API calls), and evaluation (navigation success vs.\ state verification). Our work occupies a complementary position: agents perceive heterogeneous visual inputs (ranging from natural photographs to documents, charts, and UI screenshots) shared during conversation and must translate them into high-level API tool calls within a stateful, multi-domain environment, testing the perception-to-tool-invocation pipeline rather than pixel-level UI manipulation.

A feature comparison with prior benchmarks is in Table~\ref{tab:comparison}.

\begin{table}[t]
\centering
\caption{
Comparison of representative tool-use agent benchmarks and environments. 
\textbf{Visual} indicates whether image or UI observations are part of the input. 
\textbf{Scenario} distinguishes single-turn tasks, where the user provides a complete goal upfront and the agent executes autonomously, from multi-turn tasks, where the agent must interact with the user to resolve ambiguity, obtain missing information, or handle user updates. 
\textbf{Env.} describes the execution substrate, while \textbf{Exec.} specifies the agent interface: Tool denotes structured function/tool calling, Code denotes executable code generation in a runtime such as a Python environment, and UI+Tool denotes hybrid interaction through both graphical interfaces and external tools.
}
\label{tab:comparison}
\small
\setlength{\tabcolsep}{3pt}
\begin{tabular}{lccllll}
\toprule
\textbf{Benchmark} & \textbf{Visual} & \textbf{Scenario} & \textbf{Environment} & \textbf{Execution} & \textbf{Apps/Domains} & \textbf{Tools} \\
\midrule

AppWorld~\citep{trivedi2024appworld}
    & \xmark
    & single-turn
    & API sim.
    & Code
    & mobile apps (9)
    & 457 \\

Toolathlon~\citep{li2025toolathlon}
    & \xmark
    & single-turn
    & MCP/server
    & Tool
    & software apps (32)
    & 604 \\

MCP-Atlas~\citep{bandi2026mcpatlas}
    & \xmark
    & single-turn
    & MCP/server
    & Tool
    & mixed domains (36)
    & 220 \\

ToolSandbox~\citep{lu2025toolsandbox}
    & \xmark
    & multi-turn
    & API sim.
    & Tool
    & mobile apps (11)
    & 34 \\

Gaia2~\citep{froger2026gaia}
    & \xmark
    & multi-turn
    & API sim.
    & Tool
    & service apps
    & 101 \\

ASTRA-bench~\citep{xiu2026astrabenchevaluatingtooluseagent}
    & \xmark
    & multi-turn
    & API sim.
    & Tool
    & mobile apps
    & 51 \\

$\tau^2$-Bench~\citep{barres2025tau2benchevaluatingconversationalagents}
    & \xmark
    & multi-turn
    & API sim.
    & Tool
    & telecom support
    & 30 \\

\midrule

GTA~\citep{wang2024gta}
    & \cmark
    & single-turn
    & tool suite
    & Tool
    & vision/web tasks
    & 14 \\

VisualToolBench~\citep{guo2025seeingevaluatingmultimodalllms}
    & \cmark
    & single+multi
    & tool suite
    & Tool
    & vision tasks
    & 6 \\

TIR-Bench~\citep{li2025tirbenchcomprehensivebenchmarkagentic}
    & \cmark
    & single
    & N/A
    & Optional
    & vision tasks
    & N/A \\

Agent-X~\citep{ashraf2026agentxevaluatingdeepmultimodal}
    & \cmark
    & single-turn
    & tool suite
    & Tool
    & vision/web tasks
    & 14 \\

AgentVista~\citep{su2026agentvistaevaluatingmultimodalagents}
    & \cmark
    & single-turn
    & tool suite
    & Tool
    & vision/web tasks
    & 4 \\

OSWorld-MCP~\citep{jia2025osworld}
    & \cmark
    & single-turn
    & OS/UI
    & UI+Tool
    & desktop apps (7)
    & 158 \\

MobileWorld~\citep{kong2025mobileworldbenchmarkingautonomousmobile}
    & \cmark
    & multi-turn
    & OS/UI
    & UI+Tool
    & mobile apps
    & 64 \\

\textbf{\benchname{}}
    & \cmark
    & multi-turn
    & API sim.
    & Tool+Code
    & mobile apps (16)
    & \textbf{511} \\

\bottomrule
\end{tabular}
\end{table}

%% file: sec/3_framework.tex
\section{\benchname{}}
\label{sec:framework}
\benchname{} aims to measure visual tool-calling capability: whether an agent can understand visual context, decide which tools (or code) are needed, use visual artifacts correctly across turns, and complete tasks through grounded, multi-step interaction. 
To this end, \benchname{} inherits the lightweight, interactive, and robust design philosophy of ToolSandbox~\citep{lu2025toolsandbox}, while extending text-based tool-use environments to visual settings through explicit visual artifact handling. 
To support richer and more dynamic assistant scenarios, we further incorporate domains and tools from AppWorld~\citep{trivedi2024appworld}, resulting in $16$ simulated application domains and $511$ native Python tools. Together, these design choices make \benchname{} a generic runtime for prototyping diverse multimodal tool-calling scenarios. Figure~\ref{fig:overview} provides an overview of our system design.

\subsection{Architecture}
\label{sec:architecture}
Each episode is parameterized by a \textbf{scenario}, which specifies the initial and expected world states, available tools, user instructions, associated visual assets, and completion criteria. The interaction then unfolds among an \textbf{agent}, a \textbf{user}, and an \textbf{execution environment}. An LLM plays the simulated user, the agent acts as the task solver (under evaluation), and the environment maintains the shared world state while executing tool calls. The interactions among agent, user and environment operate over two channels. The front-end interaction between the agent and user is a natural user-assistant dialogue, while the backend interaction with the environment is mediated through typed tool calls. 

\textbf{Tool Management.} Tools are implemented as native Python functions with typed signatures and metadata such as role visibility and state dependencies~\citep{lu2025toolsandbox}. Because we provide over $500$ tools in \benchname{}, exposing the full registry in a single prompt is impractical. We therefore provide a tool-discovery meta-tool, \texttt{search\_tool}, which retrieves relevant tools on demand from a natural-language query over tool documentation. Retrieved tools are added to the agent's active set, while a least-recently-used (LRU) eviction policy keeps the working set bounded. To further test tool-use robustness under varying registry sizes, we curate three additional variants from the full native toolset: a medium set with $276$ tools, a compact set with $165$ tools, and a mini set with $30$ tools.

\subsection{Multimodal Interaction}
\label{sec:multimodal}
Multimodal interaction is the main focus of this work. Building on the message-passing and world-state design of ToolSandbox~\citep{lu2025toolsandbox}, we treat images as first-class visual artifacts rather than static prompt attachments. Each image is stored in a shared image database, assigned a stable identifier, and referenced by tools and messages in the same way that other environment entities are referenced. This design makes visual artifacts persistent, reusable, and traceable across turns. Image exchange is mediated through tool calls. For example, the user can provide images to the agent during the conversation via \texttt{send\_message\_with\_image}, and the agent can similarly return visual artifacts to the user when needed. We further provide visual-specific tools and other utilities for resizing, annotating, plotting, transforming images, as well as interactive UI generation; the complete list of visual tools is given in Appendix~\ref{app:vision_tools}. Note that our benchmark does not prescribe the use of any particular visual tool: cropping, resizing, rotation, and enhancement are available, but whether to invoke them is left to the agent's own judgment rather than being mandated in the system prompt. As a result, images become part of the structured interaction trajectory rather than untracked context-window inputs. An example is shown in Figure~\ref{fig:overview}.

\subsection{Execution Modes}
\label{sec:execution_modes}
\benchname{} implements two execution modes for invoking tools: free-form \textbf{Code-Execution} and structured \textbf{Tool-Use}. In \textbf{Code-Execution} mode, the model operates through standard chat completion and writes Python code in markdown-fenced blocks, following the code-as-action paradigm~\citep{trivedi2024appworld,wang2024executable}. 
The framework parses and executes the code in a persistent sandboxed interpreter, where registered tools are available as callable Python functions. This interface naturally supports complex workflows involving loops, conditionals, intermediate variables, and flexible tool composition, with the drawback of reduced structural constraints. In \textbf{Tool-Use} mode, the model operates through the provider's structured tool-calling API, as commonly supported by OpenAI, Anthropic, and Gemini~\citep{gpt5-function-calling-api,anthropic-function-calling-api,gemini-function-calling-api}. At each step, the model either responds to the user or emits a schema-constrained tool call that is executed by the environment. This interface is easy to parse, auditable, and robust to malformed outputs, but can be cumbersome for complex control flow or multi-tool composition. To support programmatic orchestration in this mode, we provide an \texttt{execute\_code} tool that invokes the same sandboxed interpreter, and code execution remains mediated by the structured tool-calling interface.

\textbf{Safety Guardrails.} 
Executing agent-generated Python code requires careful protection regardless of how the code is invoked. 
Following the spirit of AppWorld's sandboxed execution design~\citep{trivedi2024appworld}, \benchname{} protects the interpreter with three layers of guardrails: 
(1) static checks that reject disallowed imports and unsafe primitives before execution; (2) runtime restrictions that limit file access, intercept dangerous system calls, and cap memory usage; and (3) execution limits on wall-clock time and output size. 

\subsection{Evaluation}
\label{sec:evaluation}
\benchname{} evaluates scenarios through two complementary mechanisms (illustrated on the right of Figure~\ref{fig:overview}). \textbf{State-based verification} compares initial and final world-state snapshots against a scenario's expected entity changes, providing an objective, reproducible outcome signal. \textbf{Rubric-based LLM judges} assess trajectory quality for both roles: an Agent Judge evaluates whether the agent completed the task correctly, while a User Judge checks whether the simulated user behaved coherently. Together, these mechanisms cover both the outcome and the process of each scenario. Concrete metrics are defined in Section~\ref{sec:eval_protocol}.

%% file: sec/4_benchmark.tex
\begin{figure*}[h]
\centering
\includegraphics[width=\textwidth]{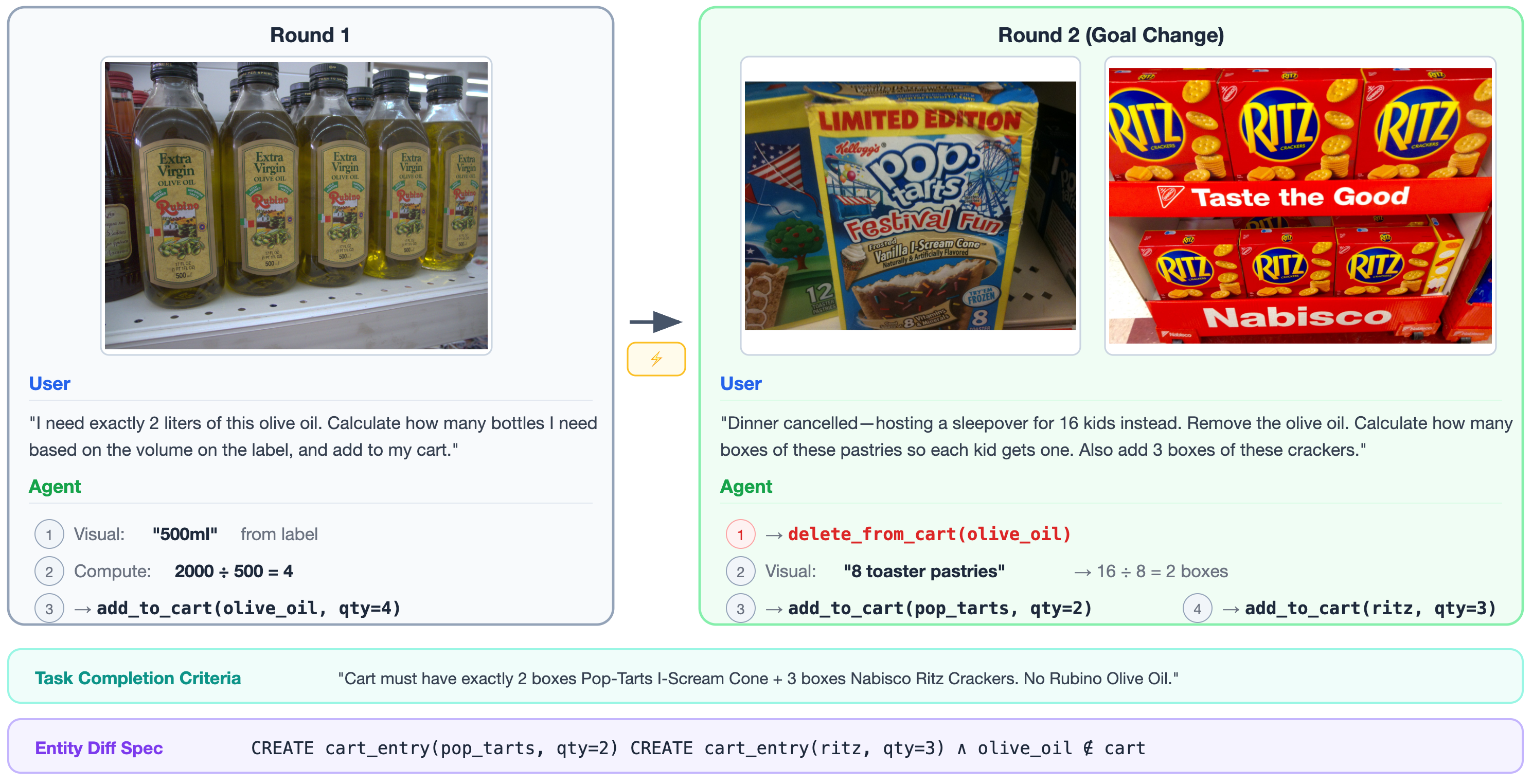}
\caption{Example scenario illustrating the interplay of benchmark design dimensions. The agent must (1)~extract numerical information from product images (visual grounding), (2)~perform arithmetic to determine quantities (compute flow type), (3)~adapt when the user changes plans mid-conversation (goal change challenge), and (4)~handle images arriving across turns (progressive arrival). Evaluation verifies the final cart state against expected entity diffs.}
\label{fig:example_scenario1}
\end{figure*}

\section{Benchmark}
\label{sec:benchmark}
The \benchname{} benchmark comprises 258 human-verified scenarios spanning diverse visual reasoning patterns, conversational challenges, and application domains. Each scenario specifies a multi-turn, multi-image task that requires the agent to extract information from visual inputs, navigate a large tool space, and produce correct state changes in a simulated device environment. We describe the design dimensions that structure the benchmark, followed by its composition and statistics. In addition, we curate $50$ variants for interactive UI applications. A scenario example is shown in Figure~\ref{fig:example_scenario1}. More examples can be found in Appendix~\ref{app:examples}.

\subsection{Design Dimensions}
\label{sec:design_dimensions}
We structure the benchmark along three orthogonal axes, namely, information flow type, challenge type, and image arrival pattern, that jointly control the reasoning pattern, conversational dynamics, and visual complexity of each scenario.

\noindent\textbf{Information flow types.}
The flow type prescribes the reasoning pattern connecting visual inputs to tool actions.
We derive seven flow types by systematically decomposing the space of operations that relate multiple information sources to action outputs, grounding our taxonomy in data processing primitives:
\emph{aggregate} (cf.\ \textsc{union}): combine information from multiple sources into a single action;
\emph{compare} (cf.\ \textsc{argmax}): evaluate alternatives against criteria to select the best;
\emph{filter} (cf.\ \textsc{where}): select a subset of items matching a condition;
\emph{compute} (cf.\ derived column): derive a new value from inputs through calculation;
\emph{validate} (cf.\ \textsc{check}): verify information against constraints before acting;
\emph{lookup chain} (cf.\ correlated subquery): resolve references sequentially, each feeding the next;
and \emph{cross-reference} (cf.\ \textsc{join}): match information across sources and act on the correspondence.
These flow types impose genuinely different reasoning demands, e.g. lookup chains require sequential retrieval from individual images, while filter and aggregate tasks require synthesizing information across multiple images simultaneously.

\noindent\textbf{Challenge types.}
We control for two dimensions of task difficulty beyond the cooperative baseline. First, two conversational challenges test agent robustness to imperfect user behavior: \emph{goal change}: the user pivots mid-conversation, requiring the agent to abandon prior progress and adapt to revised objectives; and \emph{error correction}: the user provides incorrect details (e.g., misidentifying content in an image), testing whether the agent detects discrepancies between visual evidence and user instructions rather than executing blindly. Second, state mutation controls operational complexity by requiring the agent to update or delete existing entities rather than only creating new ones, demanding entity retrieval and grounded modification instead of unconstrained creation. Scenarios without any of these properties form the cooperative baseline with create-only operations.

\noindent\textbf{Image arrival patterns.}
The arrival pattern determines when images enter the conversation:
\emph{upfront}: all images in the first message, \emph{progressive}: images distributed across turns, \emph{late}: images arriving toward the end, or \emph{mixed}: a combination of the three. Progressive delivery tests whether agents can integrate new visual context as the interaction evolves, while upfront delivery tests simultaneous multi-image working memory, which is a fundamentally different cognitive demand.

\subsection{Benchmark Composition}
\label{sec:composition}

\noindent\textbf{Image sources.}
We sourced images from the test sets of eight established vision datasets spanning complementary visual domains:
documents (DocVQA~\citep{docvqa}, OmniDocBench~\citep{ouyang2025omnidocbenchbenchmarkingdiversepdf}),
natural scene text (HierText~\citep{hiertext}),
real-world scenes (WorldVQA~\citep{zhou2026worldvqa}),
charts and plots (ChartQA-Pro~\citep{masry-etal-2025-chartqapro}, ChartMuseum~\citep{tang2025chartmuseum}),
infographics (InfographicVQA~\citep{mathew2021infographicvqa}),
and software UI screenshots (ScreenSpotPro~\citep{li2025screenspotproguigroundingprofessional}). 
This diversity ensures that benchmark performance reflects general visual perception rather than proficiency on a single image type. Given the image set, we leverage our automatic scenario generation pipeline to produce a diverse and challenging scenario set (detailed in Section~\ref{sec:scenario_generation}). Each scenario requires 3--6 images and 2--5 user interaction rounds. In total we cover 1,284 unique images over 258 scenarios, and the distribution of design dimensions is shown in Figure~\ref{fig:pipeline}.

\noindent\textbf{UI mode subset.}
We additionally convert a 50-scenario subset to UI mode, where the same underlying tasks must be completed through visual interface interaction rather than text. In this mode, the agent acts as a dynamic UI designer: it discovers backend tools, gathers information, and renders interactive screens (product cards, selection lists, confirmation dialogs) through which the user makes decisions via button clicks and selections. The key design principle is that the user's decision criterion is private, so the agent cannot complete the task without presenting options and receiving the user's choice through the UI. Entity diff specifications remain identical, ensuring that the same programmatic evaluation applies regardless of interaction modality. A third judge (UI judge) additionally evaluates whether the agent rendered appropriate interactive interfaces with correct affordances. A converted UI scenario is shown in Figure \ref{fig:UI_example}.

\begin{figure*}[t]
\centering
\includegraphics[width=\textwidth]{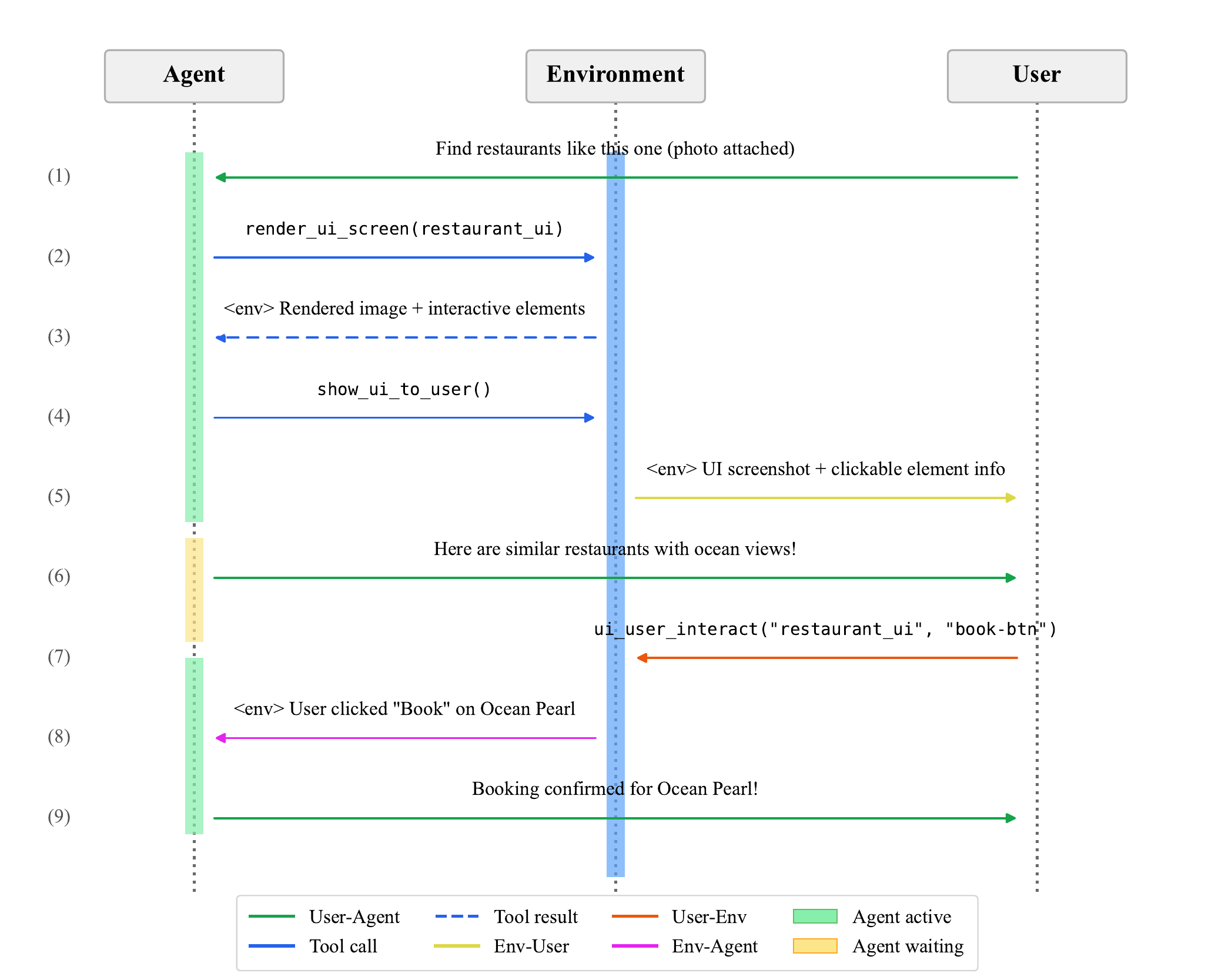}
\caption{An example information flow for an interactive UI scenario. The user simulator and agent interact through natural language on the front end, while both interact with the UI server environment through dedicated UI tools on the back end. The agent generates a new UI screen by writing Python code following the A2UI protocol, which is executed through \texttt{render\_ui\_screen}. The environment renders the resulting UI screen and returns it to the user role as an environment message. The user simulator then observes the screen and interacts with it through \texttt{ui\_user\_interact}, emulating common UI actions such as clicking and typing.}
\label{fig:ui_sequence}
\end{figure*}
To support interactive UI scenarios, we implement a set of dedicated UI tools that mediate structured visual interactions between the agent, the user, and the environment acting as a UI state server (Figure~\ref{fig:ui_sequence}). On the frontend, the agent and the user continue to communicate in natural language as in any other scenario. On the backend, the agent constructs a declarative UI description and calls a rendering tool; the environment validates, renders, and stores the resulting screen, then delivers it to the user together with metadata describing the available interactive elements. The user can then perform actions such as clicking a button or filling a form field through a dedicated interaction tool, and the environment resolves the action into structured data that is forwarded back to the agent as a notification. This design keeps the conversational surface natural and readable while making every UI state change auditable through typed tool calls. UI tools are built based on the recently published A2UI protocol~\citep{a2ui}. Figure~\ref{fig:ui_sequence} demonstrates the information flow for UI interaction.

\subsection{Evaluation Protocol}
\label{sec:eval_protocol}
By default, we use GPT-5.4~\citep{openai-54} to build the user simulator, as we find that this model achieves the highest user fidelity and natural conversation compared to other frontier models. The prompts for both user and agent roles are provided in Appendix~\ref{app:agent_prompt} and Appendix~\ref{app:user_prompt}.

\noindent\textbf{Agent implementation.}
To isolate model capability from scaffolding, all models share a single fixed harness \citep{yao2023react,wang2024executable}. Unless otherwise noted, the agent runs in the pure code-execution interface (Section~\ref{sec:execution_modes}): it operates a persistent sandboxed Python interpreter, discovers tools on demand from the full $511$-tool registry via a discovery tool under an LRU working set, and is capped at $100$ interaction steps (across agent and user roles) with a per-scenario wall-clock time limit. Native thinking is enabled for models that support it. Every model receives the same agent-role system prompt (Appendix~\ref{app:agent_prompt}) and interacts with the same GPT-5.4~\citep{openai-54} user simulator, and all trajectories are scored by a Claude 4.5 Sonnet~\citep{claude-4-5-sonnet} rubric judge. We hold this harness fixed on purpose: as agent scaffolding increasingly drives end-task performance~\citep{anthropic2025harness,openai2026harness}, comparing models under a common, well-specified harness is what makes performance differences attributable to the model rather than to the surrounding system. We separately study how the interface itself affects performance in Section~\ref{sec:agent_design}.

\noindent\textbf{Evaluator.} Each scenario is scored using the mechanisms described in Section~\ref{sec:evaluation}. \textbf{Agent Success Rate (Agent SR)} is the primary metric: the agent judge receives the full conversation trajectory and tool calls with results and then evaluates five independent criteria: task completion, instruction following, tool use validity, no side effects, and information accuracy, and a scenario passes only if all five pass. The complete agent-judge rubric prompt is provided in Appendix~\ref{app:agent_judge}.
\textbf{Entity F1} provides graded partial credit by programmatically comparing actual entity state changes against expected specifications via Hungarian matching with type-aware column similarity (exact match for identifiers, ROUGE-L for text, datetime verification for timestamps; full details in Appendix~\ref{app:eval_protocol_details}). Note that the two metrics are complementary by construction and can diverge: Entity F1 asks \emph{how close is the final state?}, tolerating a misread name or one wrong field, whereas Agent SR is a conjunctive gate asking \emph{did the agent do everything correctly?}. We analyze where they disagree in Appendix~\ref{app:metric_disagreement}. \textbf{User SR} assesses whether the simulated user behaved coherently, serving as a sanity check that user quality does not confound agent evaluation (user-judge rubric in Appendix~\ref{app:user_judge}). We show in Section~\ref{app:user_sr_filtered} that User SR correlates strongly with Agent SR and that excluding user-failing scenarios does not alter model rankings, confirming that reported performance differences reflect genuine agent capability.

\noindent\textbf{Judge model selection.} We use Claude 4.5 Sonnet~\citep{claude-4-5-sonnet} as the default judge. Among the proprietary candidates we compared during judge development (Claude 4.5 Sonnet, GPT-5.4, Gemini 3.0 Flash) it attains the highest agreement with human experts while also being the most \emph{conservative}, i.e., it has the lowest false-positive rate and therefore under- rather than over-estimates Agent SR. See Appendix~\ref{app:human-agreement} for our human-judge model agreement study.

\noindent\textbf{Evaluation validity.} We validate judging fidelity against human annotation. Three expert annotators independently labelled the same $100$ trajectories, reaching substantial inter-annotator agreement (Fleiss' $\kappa = 0.782$, mean pairwise Cohen's $\kappa = 0.782$), which establishes that the labelling task itself is reliable. Against the majority-vote human label, the agent judge achieves $86\%$ agreement (Cohen's $\kappa = 0.709$, substantial) with only $2$ false positives among $56$ human-negative trajectories (FPR $=0.036$). Importantly, the judge is miscalibrated in the conservative direction: it passes $34\%$ of the sampled trajectories where the human majority passes $44\%$, so reported Agent SR should be read as a lower bound on true performance rather than an unbiased estimate. The user judge attains high raw agreement ($77\%$) though a lower $\kappa$ driven by the rarity of user-simulation failures. We further re-score every trajectory with two additional judges drawn from different model families, Gemini 3.0 Flash and the open-weight Qwen 3.5-397B-A17B, and find model rankings essentially unchanged (Spearman $\rho = 0.95$--$0.99$), with no evidence that the Sonnet judge favors Claude-family agents (Appendix~\ref{app:judge_robustness}). Our conclusions therefore do not hinge on the choice of judge. Repeating the judge on fixed trajectories and repeating full agent rollouts both yield standard deviations below $1.3\%$ (Appendix~\ref{app:variance}), well below the gaps between performance tiers. Together with the strong Agent-SR/User-SR correlation ($r=0.81$, Figure~\ref{fig:user_sr_correlation}), this indicates that both the user simulator and the judges are reliable.

%% file: sec/5_scenario_generation.tex
\section{Scenario Generation Pipeline}
\label{sec:scenario_generation}
\begin{figure*}[t]
\centering
\includegraphics[width=\textwidth]{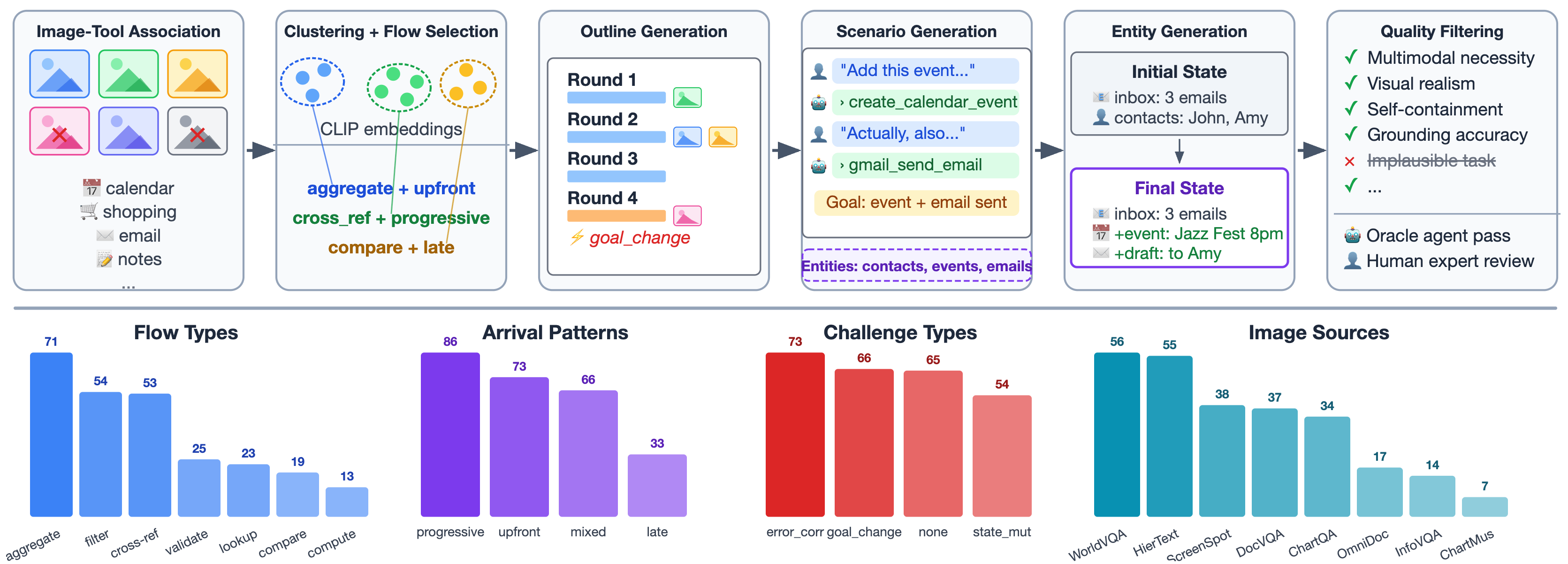}
\caption{Overview of the scenario generation pipeline (top) and the resulting benchmark distribution (bottom). \textbf{Top:} Each candidate image is first associated with relevant application domains and tools (Stage~1), then grouped via CLIP-based clustering and screened for viable information-flow types and image-arrival patterns (Stage~2). A structured multi-round outline is sampled from the viable patterns and assigned a challenge type (Stage~3), then expanded into a full multi-turn dialogue with deictic visual references and tool calls (Stage~4). Stage~5 instantiates the initial environment state together with the expected final state for deterministic state-based evaluation. Stage~6 applies LLM-based quality criteria, followed by an oracle agent solvability check and human expert review. \textbf{Bottom:} Distribution of the $258$ final scenarios across the three design dimensions of Section~\ref{sec:design_dimensions}, flow types, image-arrival patterns, and challenge types, together with the originating image-source datasets.}
\label{fig:pipeline}
\end{figure*}

We design an automated pipeline that generates multi-image, multi-turn scenarios at scale while satisfying the design dimensions defined in Section~\ref{sec:design_dimensions}.
The pipeline operates in six stages (Figure~\ref{fig:pipeline}), producing the benchmark described in Section~\ref{sec:benchmark}. All LLM-based stages use Gemini-3.1-Pro~\citep{gemini-3-1-pro}, which we selected in early pilot studies for its strong multimodal understanding and, critically, its reliability at emitting schema-conformant environment entities, a prerequisite for the complex entity staging required in Stage~5. We probe the effect of this choice with a generator ablation in Appendix~\ref{app:generator_bias}.

\noindent\textbf{Stage 1: Image--tool association.}
Not all images naturally motivate tool use.
Given an image sourced from the image corpus, an LLM identifies whether it contains genuinely actionable content and maps it to relevant application domains and specific tool functions.
Strict filtering ensures only images requiring both visual extraction and tool invocation, where the task cannot be accomplished without the image, pass to downstream stages.

\noindent\textbf{Stage 2: Image clustering and flow type selection.}
Randomly grouping images produces forced, artificial scenarios. We instead first group images by associated domains and then cluster them into groups of 3-6 using CLIP embedding similarity, ensuring high relevance within each group.
For each cluster, we assess the feasibility of each information flow type and image arrival pattern (defined in Section~\ref{sec:design_dimensions}).
A lightweight viability assessment scores each flow type and each arrival pattern for the cluster (0-3: impossible to ideal); only natural fits (score $\geq 2$) are retained.
Without this pre-selection, we observe that the LLM defaults to the simplest viable pattern, collapsing reasoning diversity.

\noindent\textbf{Stage 3: Outline generation.}
Before generating the full scenario, the pipeline produces a structured outline by selecting from the acceptable flow types and image arrival patterns identified in Stage~2, with weighted sampling that steers toward a balanced distribution across the benchmark.
The outline specifies the number of user rounds (2--5), the selected flow type and arrival pattern, required tools, and a challenge type (Section~\ref{sec:design_dimensions}).

\noindent\textbf{Stage 4: Scenario generation.}
The outline is expanded into a full scenario using flow-type-specific prompt templates that encode the assigned information flow structure, ensuring genuine multi-step reasoning rather than independent parallel actions.
The output includes per-turn user queries with deictic references to visual content, transition conditions governing when new images or corrections are introduced, a goal with all references resolved to concrete values for automated evaluation, and \emph{preexisting entity requirements}, the data that must exist in the environment for the scenario to be valid (e.g., existing calendar events, contacts, or email threads).
A user profile sampled from a pool of synthetic personas provides realistic social context for scenarios involving personalized information.

\noindent\textbf{Stage 5: Entity generation.}
Stateful evaluation requires fully specified environment state.
Using the preexisting entity requirements from Stage~4, an LLM instantiates \emph{initial entities} (concrete data populating the env so the scenario is valid) and \emph{final entities} (the expected env state after successful completion, including created, modified, and deleted entities), enabling deterministic evaluation by diffing actual post-execution state against the expected outcome.

\noindent\textbf{Stage 6: Automated quality filtering.}
Each scenario is evaluated by an LLM critic against seven quality criteria: multimodal necessity, visual realism, self-containment, conversation step integrity, goal flexibility, visual grounding accuracy, and scenario plausibility.
Scenarios passing all criteria then undergo balanced downsampling to maintain diversity across design dimensions of Section~\ref{sec:design_dimensions}. The downsampled set is then validated for solvability and entity correctness through the verification protocol described in Section~\ref{sec:quality_verification}.

\subsection{Quality Verification}
\label{sec:quality_verification}
To verify that every released scenario is executable end-to-end, not merely well-formed, we add a final layer on top of Stage~6 filtering, combining an oracle rollout check and human expert review.

\noindent\textbf{Oracle solvability check.}
For every scenario, we run an oracle agent whose system prompt is augmented with ground-truth information: the expected entity state changes, the task completion criteria, and the list of expected tools. We default to Claude 4.5 Opus as the oracle backbone; using a strong model with full ground-truth access establishes an upper bound on scenario solvability that is unconfounded by model capability. A
scenario passes only when the oracle achieves \emph{both} a near-perfect entity-diff match (Entity F1 $\geq 0.9$) and a passing Agent Judge verdict; failures are routed for correction or removal.

\noindent\textbf{Human expert review.}
Because the oracle check cannot detect scenarios that are technically solvable yet semantically misaligned with the user's intent, every scenario passing the oracle is further inspected by human expert reviewers. Verification was performed by three expert annotators who partitioned the post-downsampling pool, reviewing $100$ scenarios each. For each scenario, the annotator inspected the task specification, the attached images, the oracle-agent trajectory, the judge verdict, and the resulting state changes, then (i)~confirmed that oracle-passing scenarios were true positives rather than accidental passes, and (ii)~diagnosed oracle-failing scenarios as either problematic specifications or minor mismatches. The reviewers verify that the dialogue, expected tools, entity specifications, and challenge type are mutually consistent, and that visual references resolve unambiguously to the attached images.
Scenarios with a localized issue were corrected in place, most commonly by adding a missing staging entity, and re-run through the oracle to confirm solvability; scenarios that were unsolvable or infeasible to fix were discarded. Of the $300$ scenarios entering this stage, approximately $50$ were corrected in place and $42$ were removed, yielding the final $258$.

%% file: sec/6_experiments.tex
\begin{table*}[t]
\centering
\caption{Main results. Unless otherwise specified, all models are evaluated with thinking mode enabled using the Code-Execution interface.
}
\label{tab:main_results}
\small
\begin{tabular}{lccc}
\toprule
\textbf{Model} & \textbf{Agent Success Rate} $\uparrow$ & \textbf{Entity F1} $\uparrow$ & \textbf{Avg. Steps} $\downarrow$ \\
\midrule
Qwen 3.5-4B~\citep{qwen35blog} & 0.116 & 0.641 & 54.5 \\
Qwen 3.5-9B~\citep{qwen35blog} & 0.190 & 0.693 & 53.8 \\
Qwen 3.5-27B~\citep{qwen35blog} & 0.349 & 0.792 & 46.9 \\
Qwen 3.5-35B-A3B~\citep{qwen35blog} & 0.260 & 0.751 & 48.0 \\
Qwen 3.5-397B-A17B~\citep{qwen35blog} & 0.353 & 0.793 & 50.4 \\
GLM 4.6V~\citep{vteam2025glm45vglm41vthinkingversatilemultimodal} & 0.190 & 0.709 & 46.6 \\
KIMI 2.6~\citep{kimi2-6-tech} & 0.415 & 0.817 & 48.1 \\
\midrule
GPT-5.4 (no thinking)~\citep{openai-54} & 0.291 & 0.794 & 27.8 \\
GPT-5.4 (thinking-medium)~\citep{openai-54} & 0.361 & 0.811 & \textbf{27.6} \\
GPT-5.4 (thinking-high)~\citep{openai-54} & 0.415 & \textbf{0.865} & 36.1 \\
Gemini 3.0 Flash~\citep{gemini-3-flash} & 0.364 & 0.619 & 34.9 \\ 
Gemini 3.1 Pro~\citep{gemini-3-1-pro} & 0.481 & 0.847 & 42.6  \\
Claude 4.5 Sonnet~\citep{claude-4-5-sonnet} & 0.337 & 0.817 & 50.8 \\
Claude 4.5 Opus~\citep{claude-45-opus} & \textbf{0.488} & 0.848 & 40.8 \\
\bottomrule
\end{tabular}
\end{table*}

\begin{figure}[!t]
\centering

\begin{subfigure}[t]{0.24\linewidth}
    \vspace{0pt}
    \centering
    \includegraphics[width=\linewidth]{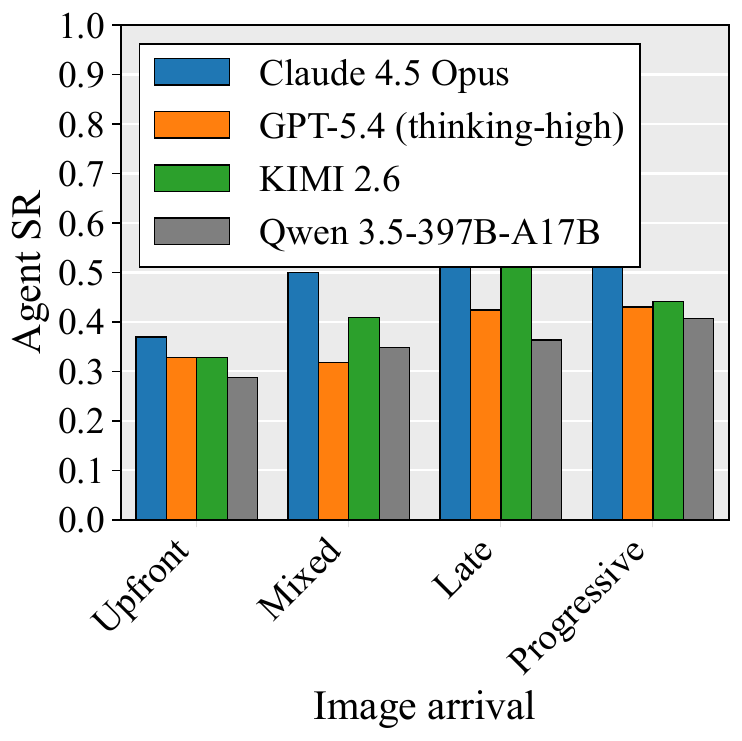}
    \caption{}
    \label{fig:image_arrival}
\end{subfigure}
\hfill
\begin{subfigure}[t]{0.24\linewidth}
    \vspace{0pt}
    \centering
    \includegraphics[width=\linewidth]{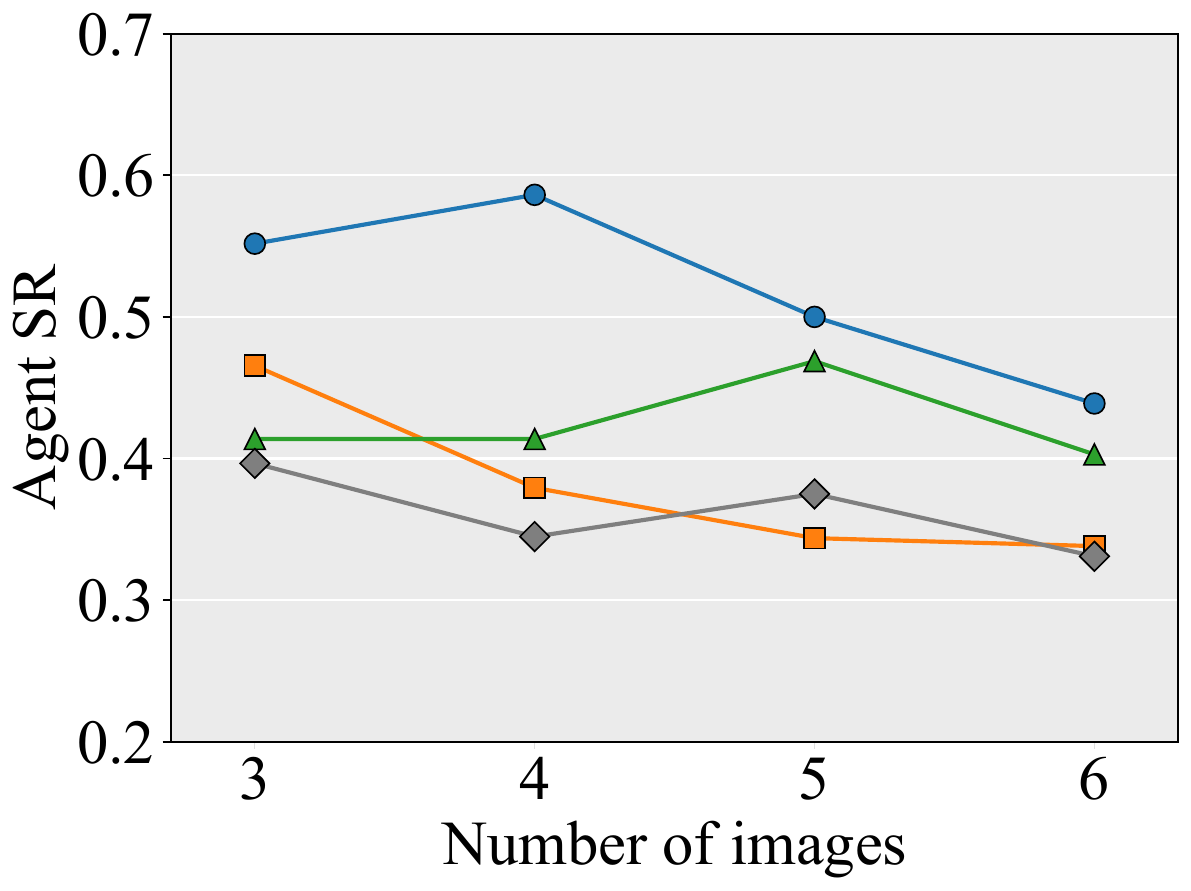}
    \vspace{0.55cm} 
    \caption{}
    \label{fig:num_images}
\end{subfigure}
\hfill
\begin{subfigure}[t]{0.24\linewidth}
    \vspace{0pt}
    \centering
    \includegraphics[width=\linewidth]{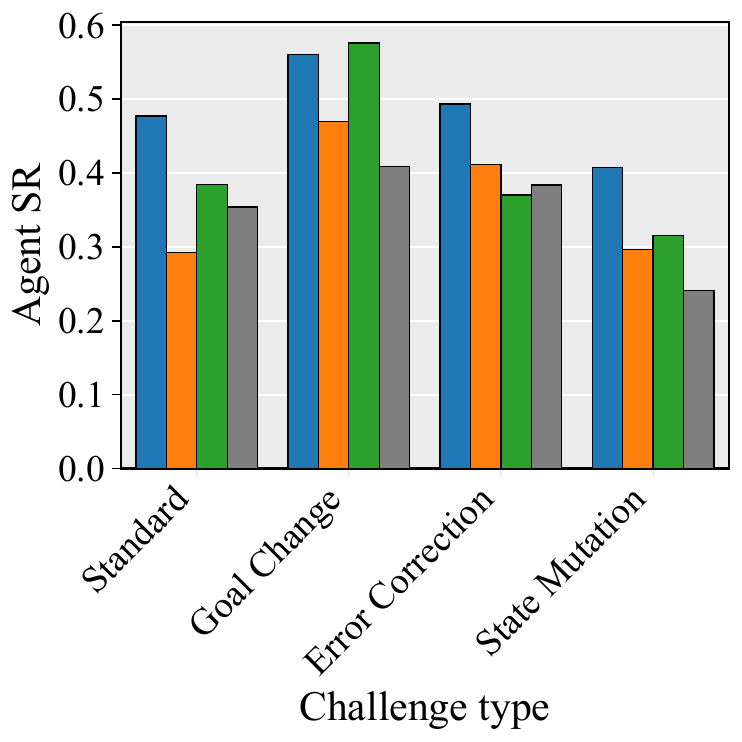}
    \caption{}
    \label{fig:challenge_type}
\end{subfigure}
\hfill
\begin{subfigure}[t]{0.24\linewidth}
    \vspace{0pt}
    \centering
    \includegraphics[width=\linewidth]{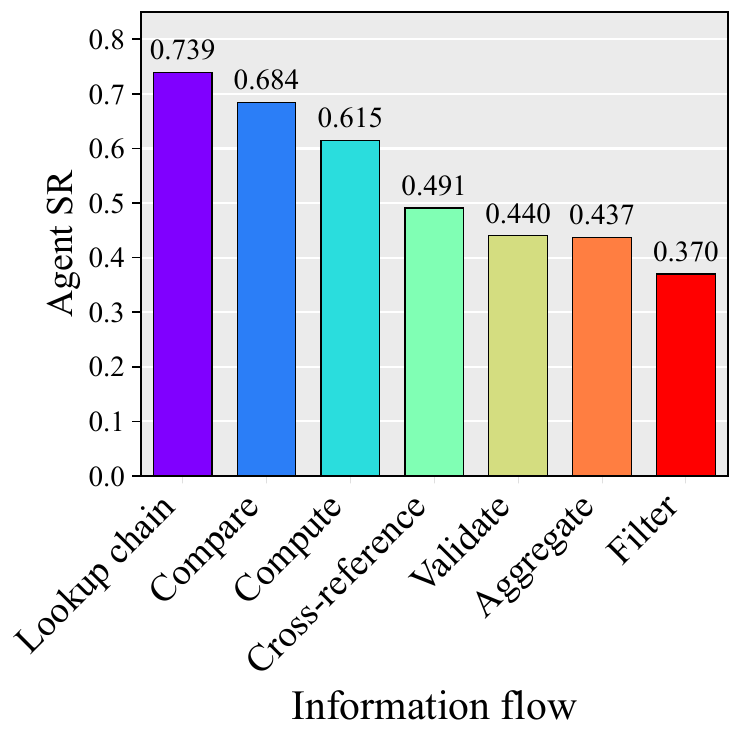}
    \caption{}
    \label{fig:agent_flow}
\end{subfigure}

\caption{
Scenario analysis.
(a) Image arrival study.
(b) Number of images within a scenario.
(c) Challenge type.
(d) Information-flow study on the Claude 4.5 Opus model.
}
\label{fig:scenario_study_images}

\end{figure}

\section{Experiments}
We evaluated $12$ strong multimodal foundation models on our \benchname{} benchmark, spanning proprietary models (GPT-5.4~\citep{openai-54}, Gemini-3.1~\citep{gemini-3-flash, gemini-3-1-pro}, Claude 4.5~\citep{claude-45-opus, claude-4-5-sonnet}) and open-weight models across a wide parameter range (Qwen 3.5 $4$B - $397$B parameters~\citep{qwen35blog}, GLM 4.6V with $106$B parameters~\citep{vteam2025glm45vglm41vthinkingversatilemultimodal}, and KIMI 2.6 with over $1$T parameters~\citep{kimi2-6-tech}). Unless mentioned otherwise, all models are evaluated using the Code-Execution interface, with the full $511$ toolset and a maximum of $100$ steps (incl. agent and user roles). The metrics are defined in Section~\ref{sec:eval_protocol}, including the Agent Success Rates (SR), Entity F1 for entity verification, and Average Steps.

\subsection{Main Results}
The main experimental results are presented in Table~\ref{tab:main_results}. We highlight four observations: \textbf{1. Visual tool calling remains challenging for frontier models.} Even the strongest model, Claude 4.5 Opus, achieves only $48.8\%$ Agent Success Rate, while GPT-5.4 (thinking-high), which obtains the highest Entity F1, still reaches only $0.865$. These results indicate that multi-step visual tool calling remains difficult, requiring image-grounded reasoning, state updates, and interaction with dynamic user behavior. \textbf{2. Agent SR, Entity F1, and Avg. Steps capture different aspects of performance.} Claude 4.5 Opus outperforms GPT-5.4 (thinking-medium) by over $12\%$ Agent SR, but reaches similar Entity F1 scores with only $3\%$ difference, reflecting the difference between process-level task completion and final-state consistency. Avg. Steps provides a further complementary signal: more steps do not necessarily lead to higher success rates, as shown by the Qwen 3.5 model series. \textbf{3. Reasoning improves task completion significantly.} 
GPT-5.4 shows a clear gain when native thinking is enabled: Agent Success Rate improves from $29.1\%$ without thinking to $36.1\%$ with medium thinking and $41.5\%$ with high thinking. Entity F1 also increases from $0.794$ to $0.865$, suggesting that additional reasoning budget is important for visual tool calling. \textbf{4. Scaling model size improves task completion significantly.}  The Qwen 3.5 family provides a controlled comparison across model scales. Performance improves substantially from $4$B to $27$B, with more than a $23$ percentage-point gain in Agent SR (from $0.116$ to $0.349$).

\subsection{Scenario Distribution Analysis}
We further study how models behave under the visual complexity and information-flow patterns designed in \benchname{} (Section~\ref{sec:benchmark}). Figure~\ref{fig:image_arrival} shows that scenarios where all images are provided upfront in the conversation are substantially harder than those with progressive image delivery, with nearly a $20\%$ success-rate gap. This is not surprising because for Upfront scenarios, later rounds often require referring back to images sent in the beginning of the conversation, when dozens of tool-call exchanges have passed, which stress tests models' long-context visual referring capability. Late and Progressive patterns, on the other hand, mostly deliver the image-relevant instructions at the same time the images arrive, making it easier for the model to pinpoint the relevant visual cues. This result suggests that \textbf{multi-image working memory is a key bottleneck for current models.}
Figure~\ref{fig:num_images} further examines model performance as the number of images increases. The success rates of Qwen 3.5, Claude 4.5 Opus, and GPT-5.4 consistently decrease with more images, highlighting the challenge of multi-image interaction. KIMI 2.6, in contrast, remains relatively stable, suggesting stronger robustness to increasing visual context. Figure~\ref{fig:challenge_type} compares performance among different challenge types. Interestingly, the goal change category yields the highest SR, because completion criteria only capture what needs to happen after the goal change, essentially reducing the task length/complexity for the agent. State Mutation, on the other hand, presents a genuine challenge to the agent, where it needs to find the necessary entities in the env before acting. Finally, Figure~\ref{fig:agent_flow} reports Agent SR for Claude 4.5 Opus across different information-flow types. 
Flow types requiring multi-image synthesis, such as filtering and aggregation, are nearly twice as difficult as single-image retrieval patterns such as lookup chains, again highlighting the need for improvement in models' visual reasoning ability. 

\noindent\textbf{Image source analysis.} We further break performance down by the visual character of the source images, grouping the eight source datasets into four categories: Natural (text-rich), Natural (less text), Docs/Charts/Tables, and UI. The full per-model table and grouping is shown in Appendix~\ref{app:image_source}. Documents, charts, and tables are the easiest category across all six models evaluated (e.g., $0.554$ for Claude 4.5 Opus), while UI screenshots are consistently the hardest ($0.372$ for Opus, and only $0.023$ for GLM 4.6V), which is consistent with the dense, small-text nature of professional UI captures. Models also differ in which visual regime suits them: Claude 4.5 Opus is substantially better on text-rich natural images than on natural images with little text ($0.48$ vs.\ $0.338$), whereas that gap is much smaller for KIMI 2.6 and Claude 4.5 Sonnet, indicating that visual bottlenecks are not uniform across models.

\subsection{Agent Design Analysis}
\label{sec:agent_design}
As motivated in Section~\ref{sec:eval_protocol}, the harness built around a model increasingly shapes its end-task performance. Having held this harness fixed for our main comparison, we now study the agent implementation design directly.

\noindent \textbf{User Simulation Study.} 
As our benchmark focuses on multi-round user-agent interaction, the quality of user simulation is highly important. To verify that user simulation quality does not confound our main findings, we recompute Agent SR and Entity F1 after excluding all scenarios where the user simulator is judged to have failed (See Appendix Table~\ref{tab:filtered_results}). Despite some variations, the overall ranking is fully preserved, showing that \textbf{the user simulation quality does not confound results.}

\begin{figure}[!t]
\centering
\begin{subfigure}[t]{0.32\linewidth}
    \centering    \includegraphics[width=\linewidth]{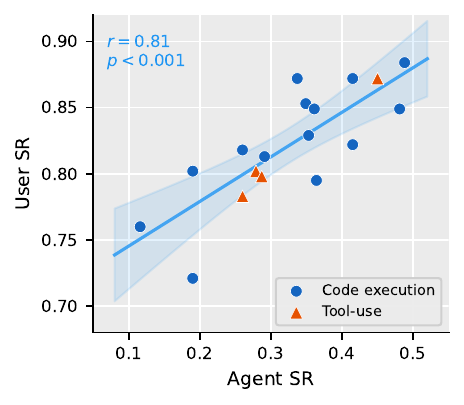}\caption{}~\label{fig:user_sr_correlation}
\end{subfigure}
\hfill
\begin{subfigure}[t]{0.32\linewidth}
    \centering
    \includegraphics[width=\linewidth]{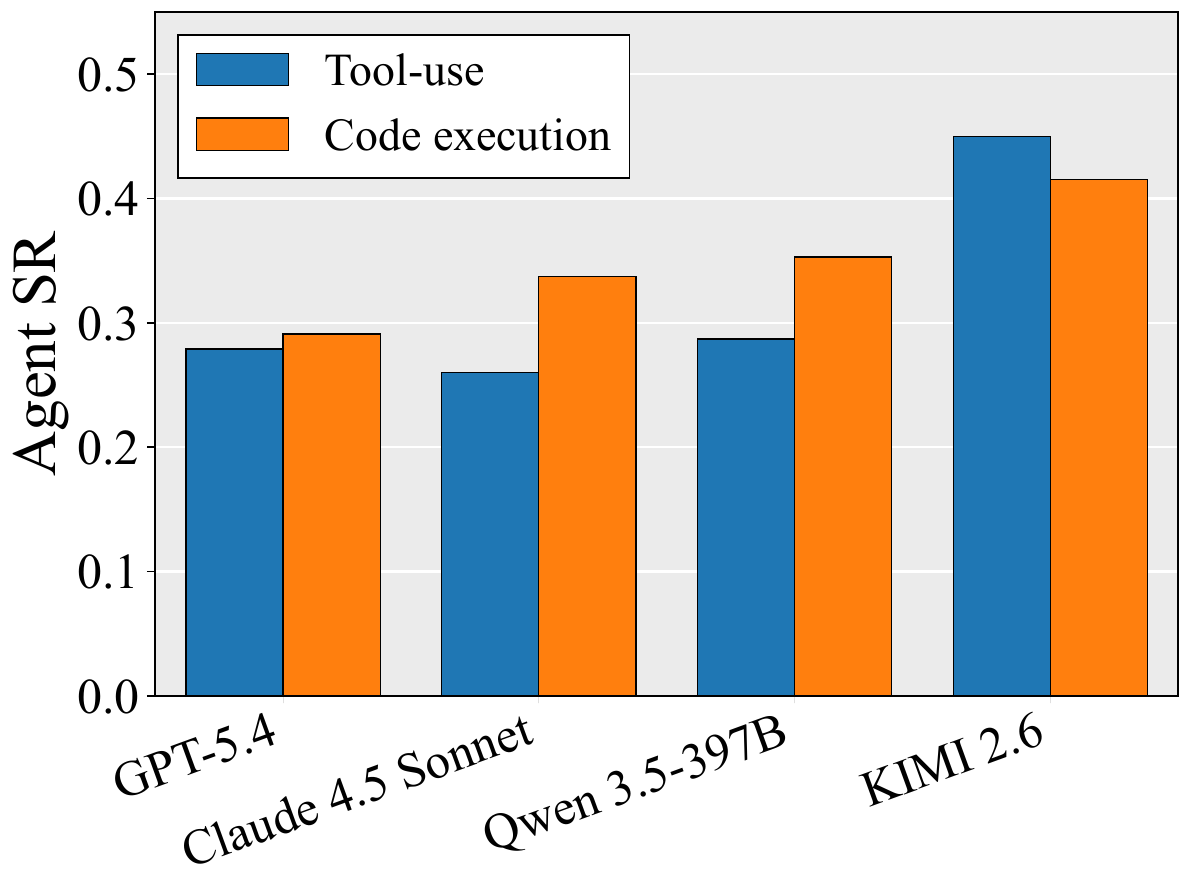}\caption{}~\label{fig:mode_comparison}
\end{subfigure}
\hfill
\begin{subfigure}[t]{0.32\linewidth}
    \centering
    \includegraphics[width=\linewidth]{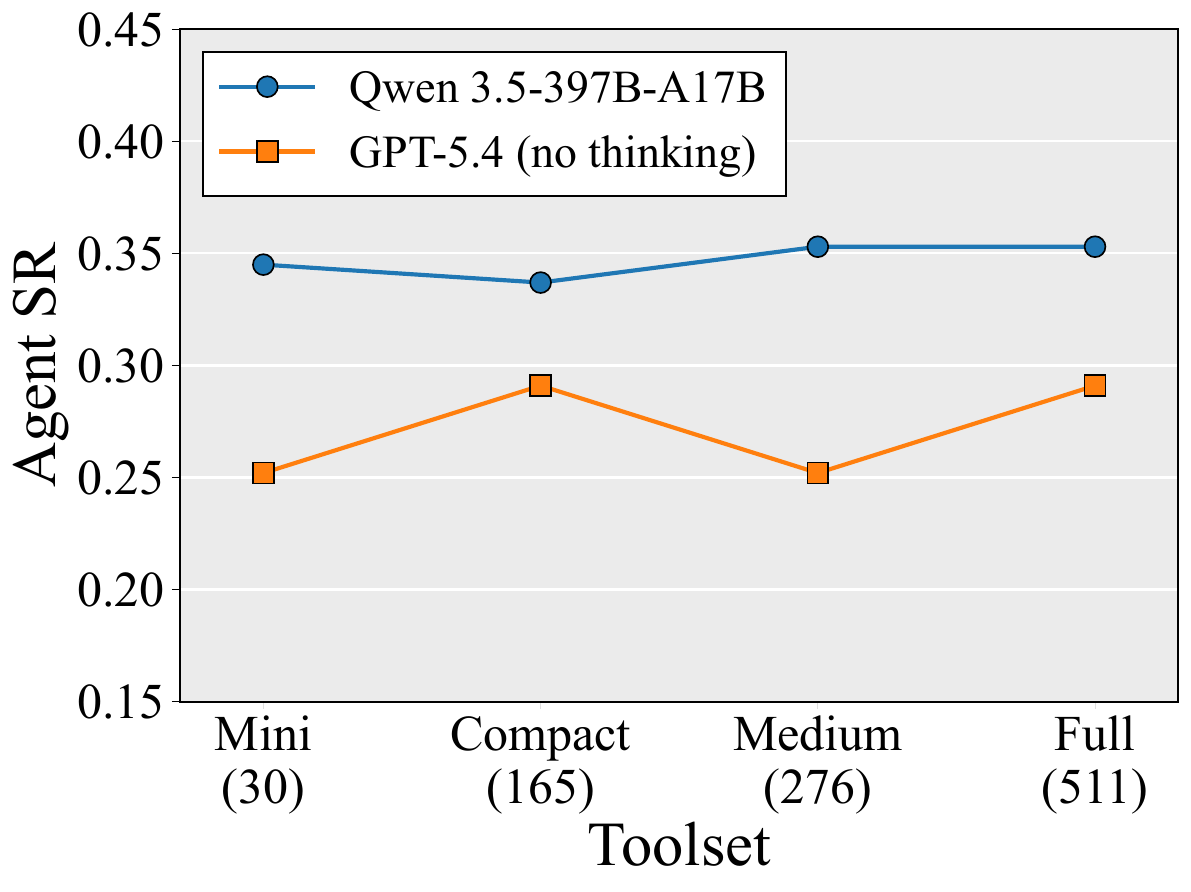}\caption{}\label{fig:toolset_comparison}
\end{subfigure}
\caption{(a). The correlation between Agent SR and User SR. (b). Comparing Tool-Use and Code-Execution interface design. (c). Agent SR with different toolset interface design.}
\label{fig:agent_design_ablation}
\end{figure}

We further plot User SR against Agent SR using $18$ model rollouts in Figure~\ref{fig:user_sr_correlation} (including both code execution and tool-use variants). The strong linear relationship (with $r=0.81$) indicates that user simulation quality is largely driven by agent behavior: agents that produce incoherent actions, fail to respond to queries, or enter retry loops destabilize the user, leading to lower User SR. This confirms that performance differences in Table~\ref{tab:main_results} reflect genuine agent capability rather than user simulation artifacts.

\noindent \textbf{Tool Discovery Analysis.}
Since agents must locate tools in a $511$-tool registry through \texttt{search\_tool}, we ask whether discovery is itself a bottleneck. We measure surfaced recall (the fraction of tools on the scenario's reference solution path that the agent discovers) and invoked recall (the fraction it actually calls), reported in Appendix~\ref{app:tool_discovery}. Although \texttt{search\_tool} currently uses simple keyword matching, it suffices as a baseline: the three strongest models surface $87$--$88\%$ of expected tools and invoke $81$--$83\%$. Critically, Claude 4.5 Sonnet and Opus achieve nearly identical tool recall yet differ by $\sim$$15\%$ in Agent SR, indicating that the tool discovery is not the limiting factor for strong models. GLM 4.6V shows a clearer discovery deficit ($70.4$ surfaced), suggesting improved tool retrieval would benefit weaker models most..

\noindent \textbf{Execution Mode Ablations}
\label{app:mode_ablation}
As mentioned in Section~\ref{sec:framework}, \benchname{} enables Tool-Use and Code-Execution modes. Here we study how models behave across execution interfaces. The results are shown in Figure~\ref{fig:mode_comparison}. Comparing Code-Execution vs.\ Tool-Use modes, we find model-dependent preferences: Claude 4.5 Sonnet and Qwen 3.5-397B strongly prefer Code-Execution mode ($+7.7$ and $+6.6$), GPT-5.4 is mode-agnostic ($+1.2$ ), and KIMI 2.6 instead shows stronger results in Tool-Use mode compared to Code-Execution. There is no conclusive evidence showing one interface is better than the other. The result is model dependent.

\noindent \textbf{Toolbox Granularity Ablation}
\label{app:toolbox_ablation}
We ablate the toolbox interface, varying from the full $511$-tool registry down to $30$ mega-tools while keeping the underlying functionality identical, all levels dispatch to the same backend APIs and produce the same state changes. The results are shown in Figure~\ref{fig:toolset_comparison}. Agent SR across toolbox levels oscillate within a small range and no clear pattern emerges. In terms of efficiency, Qwen exhibits a U-shaped step curve: Compact (165 tools) is most efficient at 45.5 steps, Full and Medium are intermediate at 50.4 and 50.3 respectively, and Mini is slowest at 61.4. GPT-5.4 shows the opposite: steps decrease monotonically from Full (27.8), Medium (26.7), Compact (26.5) to Mini (23.3). This divergence likely reflects architectural differences in how models handle tool schema complexity versus search cost.

\subsection{Failure Analysis}
\label{sec:failure_analysis}
Beyond aggregate metrics, we perform a detailed failure analysis on the strong models to understand why agents fail and where capability improvements are most needed. Our LLM rubric judge evaluates five independent criteria per scenario, namely, task completion, instruction following, tool-use validity, no side-effects, and information accuracy. A scenario passes only if all five criteria pass. By examining which criteria fail and how they co-occur, we construct a root-cause taxonomy of failures. \textit{Factual error}: task completion passes but information accuracy fails. The agent did the right workflow but extracted wrong information from images (e.g. misidentified an object, read wrong value from image), \textit{Task failure}: the agent couldn't complete the primary action at all - missing entire workflow steps, can't find resources, or enters unrecoverable loops, which is a planning/execution failure. \textit{Incomplete execution}: Both task completion and information accuracy pass, but instruction following fails. The agent completed the task with correct facts but missed specific requirements, e.g. omitted one of N items. \textit{Uncontrolled behavior}: Only no side effects fails. The agent performed unintended actions beyond the task scope, e.g. creating extra entities, leaving residual state from earlier attempts etc.

\begin{table}[t!]
\centering
\caption{Failure root-cause taxonomy. Percentages are relative to each model's total failures.}
\label{tab:failure_taxonomy}
\small
\resizebox{\linewidth}{!}{%
\begin{tabular}{lllll}
\toprule
\textbf{Model} & \textbf{Factual Error} & \textbf{Task Failure} & \textbf{Incomplete Exec.} & \textbf{Uncontrolled Behav.} \\
\midrule
Claude 4.5 Opus~\citep{claude-45-opus} (132) & 70 (53.0\%) & 31 (23.5\%) & 19 (14.4\%) & 12 (9.1\%) \\
GPT-5.4 (thinking)~\citep{openai-54} (151) & 76 (50.3\%) & 35 (23.2\%) & 20 (13.2\%) & 20 (13.2\%) \\
KIMI 2.6~\citep{kimi2-6-tech} (151) & 74 (49.0\%) & 51 (33.8\%) & 14 (9.3\%) & 12 (7.9\%) \\
Qwen 3.5-397B-A17B~\citep{qwen35blog} (167) & 85 (50.9\%) & 53 (31.7\%) & 12 (7.2\%) & 17 (10.2\%) \\
\bottomrule
\end{tabular}%
}
\end{table}

\begin{figure*}[t]
\centering
\includegraphics[width=0.49\textwidth]{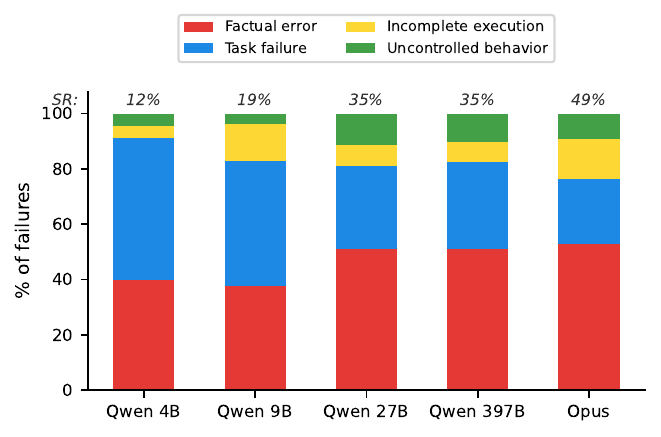}
    \caption{Failure root-cause distribution across model scale. As models grow, task failures (planning) give way to factual errors (visual precision) as the dominant failure mode.}
    \label{fig:failure_scaling}
\end{figure*}

Table~\ref{tab:failure_taxonomy} reveals that half of failures for all models stem from factual errors, cases where the agent successfully completes the task workflow and uses tools correctly, but extracts incorrect information from the visual input. For example, sub-categorizing these 70 Claude Opus failures, we find object misidentification dominates (29 cases), followed by fine-grained text extraction errors (16 cases). In rubric terms these are not planning failures: task completion and tool-use validity pass while information accuracy fails, i.e., the agent knows what to do but cannot precisely perceive what it sees. We read this as evidence that \textbf{visual precision, rather than tool-use competence, is the dominant failure mode for capable models.}

Figure~\ref{fig:failure_scaling} tracks the failure taxonomy in the Qwen-3.5 models scaling curve and Claude 4.5 Opus. We observe a cross-over between visual perception and planning errors: at 4B, task failures dominate (51\% of failures), reflecting an inability to plan and execute multi-step workflows. By 27B, task failures drop to 30\% and factual errors become the dominant mode (51\%). This crossover continues to Opus, where factual errors account for 53\% of failures and task failures only 23\%. The result suggests that for small models (below $\sim$27B parameters), the dominant failure mode is agentic planning: models cannot figure out what to do. At and above this scale, planning failures become markedly less frequent and the dominant mode shifts to visual precision: models cannot accurately perceive what they see. This suggests fundamentally different research directions for improving small vs.\ large models on visual tool-calling tasks, i.e. \textbf{model scaling reveals a planning-to-precision failure pattern crossover.}

We further show an example factual error case by Claude 4.5 Opus agent in Figure \ref{fig:opus_error} and several failure error cases in the Appendix.

\subsection{UI Mode}
\begin{table}[t]
\centering
\caption{UI mode results (50-scenario subset, Tool-Use mode). All models exhibit dramatic performance drops compared to text-based interaction on the same underlying tasks.}
\label{tab:ui_results}
\small
\begin{tabular}{lccccc}
\toprule
\textbf{Model} & \textbf{Agent SR} & \textbf{UI SR} & \textbf{User SR} & \textbf{Entity F1} & \textbf{Avg. Steps} \\
\midrule
Claude 4.5 Opus~\citep{claude-45-opus} & 0.200 & 0.280 & 0.680 & 0.609 & 74.5 \\
GPT-5.4 (thinking)~\citep{openai-54} & 0.020 & 0.360 & 0.340 & 0.279 & 92.5 \\
KIMI 2.6~\citep{kimi2-6-tech} & 0.260 & 0.420 & 0.514 & 0.699 & 81.7 \\
Qwen 3.5-397B-A17B~\citep{qwen35blog} & 0.100 & 0.200 & 0.460 & 0.619 & 75.8 \\
\bottomrule
\end{tabular}
\end{table}

As a preliminary exploration, we evaluate four models on a 50-scenario UI subset, where agents must render interactive visual interfaces for user decision-making rather than communicating through text in Table~\ref{tab:ui_results}. Even the best model (KIMI 2.6) achieves only 26.0\% Agent SR, a sharp drop from its 45.0\% on the same underlying tasks in text-based Tool-Use mode. The primary bottleneck is UI rendering itself: models frequently fail to produce functional interfaces with correct affordances, triggering retry loops that inflate average steps to $75$-$93$. These results indicate that designing and rendering interactive interfaces on-the-fly represents a substantially harder challenge than text-based tool calling, pointing to a promising direction for future benchmark expansion. In other words, \noindent\textbf{UI mode reveals a new capability frontier.}

%% file: sec/7_conclusion.tex
\section{Conclusion}

We introduced \benchname{}, a unified framework and benchmark for evaluating visually grounded tool-calling agents. The framework provides a stateful, multi-turn, multi-image simulation environment spanning $511$ tools across $16$ application domains, with both code-execution and structured tool-use interfaces. The accompanying benchmark contributes $258$ human-verified scenarios, together with $50$ interactive UI variants, produced by a scalable, information-flow-guided generation pipeline that reduces the annotation burden to targeted human review. Evaluating $12$ frontier and open-weight models, we find that visual tool calling is far from solved: even the strongest model completes fewer than half of the tasks.

Our analysis surfaces several observations that we hope will guide future work. First, visual precision, rather than tool-use competence, is the dominant failure mode for capable models: over half of the failures of the strongest models stem from misreading visual inputs despite otherwise correct task workflows. Furthermore, failure modes shift with scale, as smaller models fail primarily at planning (deciding what to do) whereas larger models fail at perception (accurately grounding what they see), implying different research directions at different capability levels. Multi-image working memory is a further key limitation, with performance degrading as more images accumulate and especially when images must be recalled from early in a long conversation. While both reasoning budget and model scale improve task completion, neither closes the visual-precision gap. Finally, the agent harness matters and its effect is model-dependent, as no single execution interface or toolbox granularity is uniformly best, underscoring the value of comparing models under a common, well-specified harness; and interactive UI generation emerges as a substantially harder frontier, with sharp performance drops relative to text-based tool use.

\section{Limitations}
\label{app:limitations}
Our work has several limitations. First, our primary metric relies on an LLM judge (Claude 4.5 Sonnet), which achieves $86\%$ agreement ($\kappa = 0.709$) with the majority label of three independent expert annotators. While this represents substantial agreement with a conservative bias, the judge may systematically miss certain failure modes. We also use Claude 4.5 Opus as the oracle solvability filter, which might inflate the performance of Claude-family models. Second, our claim that visual precision dominates strong-model failures rests on decomposing rubric judgments rather than on a controlled manipulation. Future study should have a cleaner decomposition of scenarios with and without visual inputs. On the data selection during benchmark construction, even though we draw source images exclusively from the test splits of existing public datasets, and each scenario recombines several images with newly generated user goals, tool-use workflows, and environment states. Nevertheless, image-level contamination might still exist, and future versions of our benchmark should incorporate newly captured, previously unpublished images. Finally, to keep model comparisons fair, we evaluate every model under a single fixed harness (Section~\ref{sec:eval_protocol}) rather than tuning the scaffolding per model. Since agent performance is sensitive to the harness, our reported numbers reflect model capability under this specific, standardized setup rather than each model's best achievable performance under a bespoke, model-specific harness. Systematically exploring the agent harness design space is a promising direction for future work.

%% file: sec/X_suppl.tex
\appendix

\section{Evaluation Protocol Details}
\label{app:eval_protocol_details}

\subsection{Agent Success Rate (Agent SR).}
The LLM rubric judge (Claude 4.5 Sonnet) evaluates five independent criteria, each scored pass/fail:
\begin{itemize}
    \item \textbf{Task completion}: Was the primary action performed and did the tool call succeed?
    \item \textbf{Instruction following}: Were all specific details (names, dates, amounts, recipients) correct?
    \item \textbf{Tool use validity}: Were tools used validly and sufficiently, without requiring a specific sequence?
    \item \textbf{No side effects}: Did the agent avoid unintended actions (extra writes, wrong recipients)?
    \item \textbf{Information accuracy}: Is extracted information factually correct, cross-checked against images and database state?
\end{itemize}
A scenario passes only if all five criteria pass. The judge receives the full conversation, chronological tool calls with results and all images. The judge trusts environment outputs (tool results, database state) but not the agent's natural-language claims, preventing inflated scores from models that claim success without achieving it.

\subsection{Entity F1.}
Each scenario defines expected creates, updates, and deletes with per-column similarity measures. The evaluation uses Hungarian matching to optimally align actual entities against expected ones, computing per-entity scores as the mean of column-level similarities. Column similarity is determined by data type: exact match for identifiers and numeric fields, ROUGE-L for free-text content (titles, descriptions, message bodies), and datetime verification for timestamps. The final F1 score is:
\[
\text{Entity F1} = \frac{2 \cdot \text{Precision} \cdot \text{Recall}}{\text{Precision} + \text{Recall}}
\]
where precision measures the fraction of actual changes that match expectations, and recall measures the fraction of expected changes that were realized. A guardrail check verifies that non-specified tables remain unchanged; if violated, Entity F1 is set to zero.

\subsection{User Success Rate (User SR).}
A separate LLM rubric judge (Claude 4.5 Sonnet) evaluates the user simulator, rather than the agent, on four independent criteria, each scored pass/fail:
\begin{itemize}
    \item \textbf{Request fidelity}: Did the user communicate all task-critical requests and details from its instruction, without dropping, contradicting, or redundantly re-delivering them?
    \item \textbf{Conversational naturalness}: Do the user's messages sound human, i.e., concise, casual, and free of leaked tool-call syntax?
    \item \textbf{Grounded consistency}: Are the user's facts and preferences stable across turns and grounded in its instruction or the observed conversation, with no hallucinated knowledge?
    \item \textbf{Tool channel correctness}: Were tools (e.g., image delivery) used correctly, with all required images actually delivered?
\end{itemize}
A scenario passes only if all four criteria pass. The judge treats the hidden \texttt{SYSTEM}$\rightarrow$\texttt{USER} instruction as ground truth for the user's intended behavior and evaluates only the user's conduct, explicitly not penalizing the user for the agent's execution errors, tool failures, or values the agent misread from images. It is also made aware of the scenario's challenge type (e.g., intentional errors in error-correction scenarios are by design and must not be scored as inconsistencies). The full user-judge rubric is provided in Appendix~\ref{app:user_judge}.

\section{Human Agreement Study with LLM Judges}
\label{app:human-agreement}

We quantify agreement between LLM judges and human expert annotators on a sample of trajectories.

\subsection{Study Design}

We randomly sampled $N = 100$ trajectories across five agent
configurations: Claude 4.5 Opus, Claude 4.5 Sonnet, GPT-5.4 (with and
without reasoning), and Qwen 3.5-397B-A17B (20 trajectories each).
Three expert annotators independently labelled all $100$ trajectories blind to the judge verdicts. All annotators have substantial expertise in tool-calling agent design and development.

\paragraph{Annotation protocol.}
Each annotator received exactly the information available to the LLM judge, namely the judge system prompts, the task completion criteria, and the per-trajectory entity database diffs, so that the study measures agreement on the same judging task rather than differences caused by information asymmetry. For the agent judge, annotators inspect entity-diff discrepancies, trace their causes in the trajectory, verify whether they violate the completion criteria, and check whether each user request is satisfied. For the user judge, the annotator examine the scripted goal changes and corrections and evaluate conversational naturalness, correct image delivery, and adherence to the script.

Each trajectory received two binary labels:
\begin{itemize}
  \item \textbf{Agent verdict} (\texttt{pass}/\texttt{fail}): Did the agent complete the task?
  \item \textbf{User verdict} (\texttt{pass}/\texttt{fail}): Did the user simulator behave plausibly? The rubric is lenient: mild unnaturalness, premature termination on perceived agent error, and incomplete trajectories from exhausted turn budgets are treated as \texttt{pass}.
\end{itemize}

\subsection{Inter-Annotator Agreement}
Before comparing against the LLM judges, we assess whether the labelling task itself is well-posed. Across the three independent annotations of the same $100$
agent trajectories, mean pairwise raw agreement is $0.893$ ($[0.890, 0.900]$), mean pairwise Cohen's $\kappa$ is $0.782$ ($[0.772, 0.798]$), and Fleiss'
$\kappa$ over all three annotators is $0.782$. All of these fall in the substantial agreement range, establishing that the agent-verdict labelling task is reliable and that a single annotator's results are not driving the reference labels. We therefore consolidate the three annotations into a single reference label per trajectory by majority vote; the resulting human-majority pass rate is $0.44$.

\subsection{Judge Agreement with the Human Majority Label}

Table~\ref{tab:judge_agreement} reports agreement between the human-majority
label and four candidate judge models spanning four model families.

\begin{table}[h]
\centering
\caption{Agent-judge agreement with the three-annotator human majority label on $100$ sampled trajectories ($n=100$; human-majority pass rate $=0.44$). All four judges reach substantial chance-corrected agreement.}
\label{tab:judge_agreement}
\small
\begin{tabular}{lrrrrr}
\toprule
\textbf{Judge} & \textbf{Raw} & \textbf{Cohen's} $\kappa$ & \textbf{FPR} & \textbf{FNR} & \textbf{Pass rate} \\
\midrule
Claude 4.5 Sonnet (primary)~\citep{claude-4-5-sonnet} & 0.86 & 0.709 & \textbf{0.036} & 0.273 & 0.34 \\
Qwen 3.5-397B-A17B~\citep{qwen35blog} & \textbf{0.87} & \textbf{0.741} & 0.179 & 0.068 & 0.51 \\
Gemini 3.0 Flash~\citep{gemini-3-flash} & 0.84 & 0.677 & 0.161 & 0.159 & 0.46 \\
GPT-5.4 (thinking)~\citep{openai-54} & 0.80 & 0.578 & \textbf{0.036} & 0.409 & 0.28 \\
\bottomrule
\end{tabular}
\end{table}

All four judges reach substantial chance-corrected agreement with the human
majority. Among the proprietary candidates we compared during judge development
(Claude 4.5 Sonnet, GPT-5.4, Gemini 3.0 Flash), Claude 4.5 Sonnet attains the
highest agreement, close to the inter-annotator ceiling of $0.782$, which
motivated its selection as the default judge. It is also the most
conservative judge: with the lowest false-positive rate it passes only
$2$ of the $56$ human-negative trajectories, so its pass verdicts are
highly trustworthy and reported Agent SR under-estimates rather than inflates
true performance.

The open-weight Qwen 3.5-397B-A17B judge attains comparable overall agreement
and reproduces the model ranking (Spearman $\rho = 0.951$;
Appendix~\ref{app:judge_robustness}), so it can serve as a fully reproducible,
high-fidelity alternative requiring no proprietary API. Its errors, however,
lean in the opposite direction from Sonnet's: it produces more false positives
than false negatives ($\text{FPR}=0.179$ vs.\ $\text{FNR}=0.068$). Because we
prefer a conservative judge that minimizes false positives, Sonnet remains the
default.

\begin{table}[h]
\centering
\small
\caption{Confusion matrices.}
\begin{minipage}{0.45\textwidth}
\centering
\caption*{Agent judge vs.\ human majority}
\begin{tabular}{lrrr}
\toprule
 & judge=pass & judge=fail & total \\
\midrule
human=pass & 32 & 12 & 44 \\
human=fail & 2 & 54 & 56 \\
\bottomrule
\end{tabular}
\end{minipage}
\hfill
\begin{minipage}{0.45\textwidth}
\centering
\caption*{User judge vs.\ human}
\begin{tabular}{lrrr}
\toprule
 & judge=pass & judge=fail & total \\
\midrule
human=pass & 73 & 17 & 90 \\
human=fail & 6 & 4 & 10 \\
\bottomrule
\end{tabular}
\end{minipage}
\end{table}

\paragraph{User judge.}
The user judge was labelled by a single annotator, and its class distribution is
strongly imbalanced: 90/100 trajectories are human-labelled \texttt{pass}, so
raw agreement ($77\%$) is dominated by the majority class and the low
$\kappa$ ($0.142$) reflects the difficulty of detecting the rare
user-failure class. Given the small overall proportion of user-simulation
failures, we believe the existing user simulator can act as a reliable
simulation. As an additional check, Section~\ref{app:user_sr_filtered} shows
that removing all scenarios with a failed user simulator preserves the model
ranking.

\section{Robustness of Results}
\subsection{Benchmarking Results without Failed User Simulators}
\label{app:user_sr_filtered}
\begin{table}[h]
\centering
\caption{Results after removing scenarios where the user simulator failed. N indicates the number of retained scenarios (out of 258 total). Rankings are preserved despite the filtering being proportionally more favorable to weaker models.}
\label{tab:filtered_results}
\small
\begin{tabular}{lccc}
\toprule
\textbf{Model} & \textbf{N} & \textbf{Agent SR} $\uparrow$ & \textbf{Entity F1} $\uparrow$ \\
\midrule
Qwen 3.5-4B & 196 & 0.122 & 0.662 \\
Qwen 3.5-9B & 207 & 0.203 & 0.705 \\
Qwen 3.5-27B & 220 & 0.377 & 0.807 \\
Qwen 3.5-35B-A3B & 211 & 0.280 & 0.761 \\
Qwen 3.5-397B-A17B & 214 & 0.374 & 0.805 \\
GLM 4.6V & 186 & 0.226 & 0.718 \\
KIMI 2.6 & 225 & 0.413 & 0.840 \\
\midrule
GPT-5.4 (no thinking) & 209 & 0.335 & 0.806 \\
GPT-5.4 (thinking-medium) & 219 & 0.388 & 0.815 \\
GPT-5.4 (thinking-high) & 212 & 0.458 & 0.873 \\
Gemini 3.0 Flash & 205 & 0.410 & 0.652 \\
Gemini 3.1 Pro & 219 & 0.498 & 0.831 \\
Claude 4.5 Sonnet & 225 & 0.360 & 0.833 \\
Claude 4.5 Opus & 228 & 0.504 & 0.848 \\
\bottomrule
\end{tabular}
\end{table}

\subsection{Judge Model Robustness}
\label{app:judge_robustness}

Because Agent SR is mediated by an LLM judge, we test whether our conclusions
are an artifact of the judge model. We re-scored all trajectories with
two additional judges drawn from different model families than the default:
Gemini 3.0 Flash and the open-weight Qwen 3.5-397B-A17B. All three judges use
the identical rubric prompt (Appendix~\ref{app:agent_judge}); only the judge
backbone varies. Results are in Table~\ref{tab:cross_judge}.

\begin{table}[h]
\centering
\caption{Agent SR (\%) under three judge models from three different model families. Rankings are nearly identical across judges (pairwise Spearman $\rho = 0.95$--$0.99$). GPT-5.4 refers to the thinking-high configuration. The Sonnet-judge column reproduces Table~\ref{tab:main_results}.}
\label{tab:cross_judge}
\small
\begin{tabular}{lccc}
\toprule
\textbf{Agent Model} & \textbf{Sonnet 4.5 judge} & \textbf{Gemini 3.0 Flash judge} & \textbf{Qwen 3.5-397B judge} \\
\midrule
Qwen 3.5-4B & 11.6 & 15.1 & 21.3 \\
Qwen 3.5-9B & 19.0 & 20.5 & 26.4 \\
Qwen 3.5-27B & 34.9 & 39.5 & 47.3 \\
Qwen 3.5-35B-A3B & 26.0 & 27.1 & 36.1 \\
Qwen 3.5-397B-A17B & 35.3 & 39.6 & 44.5 \\
GLM 4.6V & 19.0 & 24.2 & 24.2 \\
KIMI 2.6 & 41.5 & 50.4 & 51.1 \\
GPT-5.4 (thinking) & 41.5 & 46.4 & 47.2 \\
Claude 4.5 Sonnet & 33.7 & 36.8 & 36.8 \\
Claude 4.5 Opus & \textbf{48.8} & \textbf{53.5} & \textbf{58.5} \\
\bottomrule
\end{tabular}
\end{table}

Model rankings are nearly identical across all three judges (pairwise Spearman $\rho = 0.95$--$0.99$), with Claude 4.5 Opus strongest under every judge. We find no evidence of self-preference by the Sonnet judge: Claude 4.5 Sonnet itself ranks below several non-Claude agents, and relative to the Gemini judge the Sonnet judge favors Claude agents over non-Claude agents by only $0.35\%$ ($-3.9\%$ vs.\ $-4.2\%$). The Qwen judge, by contrast, exhibits a clear family effect, scoring Qwen-family agents $6.8\%$ higher than the Gemini judge does versus only $1.3\%$ for non-Qwen agents, which underscores that judge selection and calibration matter and motivates the human-alignment study we used to choose the default judge (Appendix~\ref{app:human-agreement}).

\subsection{Run-to-Run Variance}
\label{app:variance}
We quantify the two independent sources of stochasticity in the pipeline: the agent rollout and the LLM judge.

We repeated full agent rollouts three times end-to-end (fresh environment, fresh user simulator, fresh agent) for two models. The standard deviation of Agent SR across runs is $1.21\%$ for GPT-5.4 (thinking-high) and $0.70\%$ for KIMI 2.6. We separately ran the agent judge three times over fixed trajectories, isolating judge stochasticity from rollout stochasticity (Table~\ref{tab:judge_variance}). Standard deviations are $0.95\%$ and $0.79\%$, with ranges under $2\%$.

\begin{table}[h]
\centering
\caption{Agent SR (\%) across three independent judge runs over identical fixed trajectories, isolating LLM-judge stochasticity from agent-rollout stochasticity.}
\label{tab:judge_variance}
\small
\begin{tabular}{lcccccc}
\toprule
\textbf{Model} & \textbf{Run 1} & \textbf{Run 2} & \textbf{Run 3} & \textbf{Mean} & \textbf{SD} & \textbf{Range} \\
\midrule
GPT-5.4 (code-exec) & 41.51 & 43.40 & 42.64 & 42.52 & 0.95 & 1.89 \\
KIMI 2.6 (code-exec) & 41.50 & 40.38 & 41.89 & 41.26 & 0.79 & 1.51 \\
\bottomrule
\end{tabular}
\end{table}

From the results we conclude that the variance from both sources of noise are small, and neither changes model ranks in the repeated runs.

\section{Additional Analysis}
\subsection{Scenario Generation Pipeline Ablation}
\label{app:generator_bias}
All LLM-based stages of the generation pipeline use Gemini-3.1-Pro
(Section~\ref{sec:scenario_generation}), which raises the question of whether
the generated scenarios systematically favor Gemini-family models or otherwise
encode generator-specific structure. To isolate the generator's effect, we
re-ran scenario generation (Stage~4) and entity generation (Stage~5) with
GPT-5.4 on the same Stage-3 outlines as the official benchmark, holding
image selection, information-flow type, arrival pattern, and outline fixed so
that only the generation model varies.

\noindent\textbf{Funnel.}
Of the $258$ outlines, GPT-5.4 failed to produce schema-conformant entities for
$31\%$ ($80$) of scenarios, a failure mode from which Gemini-3.1-Pro is nearly
free. Of the $178$ scenarios that generated successfully, the Stage-6 quality
critic rejected $78$, leaving $100$ that we evaluated. Note that this ablation
subset did not pass the oracle-solvability check or human expert review
that every released scenario undergoes.

\begin{table}[h]
\centering
\caption{Generator ablation. Agent SR on $100$ scenarios generated by Gemini-3.1-Pro (the released pipeline) versus $100$ scenarios generated by GPT-5.4 from identical Stage-3 outlines. Rankings are preserved (Spearman $\rho=0.80$, Pearson $r=0.87$), and the GPT-5.4 agent does not benefit from GPT-5.4-generated scenarios. Note that the GPT-5.4-generated subset did not undergo oracle-solvability or human verification, so absolute scores are not comparable across columns.}
\label{tab:generator_bias}
\small
\begin{tabular}{lcc}
\toprule
\textbf{Agent Model} & \textbf{Gemini-generated (100)} & \textbf{GPT-5.4-generated (100)} \\
\midrule
Claude 4.5 Sonnet & 0.35 & 0.26 \\
GPT-5.4 (thinking) & 0.39 & 0.31 \\
Qwen 3.5-397B-A17B & 0.37 & 0.36 \\
Claude 4.5 Opus & \textbf{0.52} & \textbf{0.43} \\
\bottomrule
\end{tabular}
\end{table}

\noindent\textbf{Findings.}
Table~\ref{tab:generator_bias} shows that the model ranking is preserved under
the alternative generator (Spearman $\rho = 0.80$, Pearson $r = 0.87$; Claude
4.5 Opus ranked first and Claude 4.5 Sonnet last under both generators). We also
observe no generator-family advantage: the GPT-5.4 agent performs worse
on GPT-5.4-generated scenarios than on Gemini-generated ones ($0.31$ vs.\
$0.39$). We do not interpret the uniformly lower absolute scores on the
GPT-5.4-generated subset, since this subset skipped oracle filtering and human
verification and therefore likely contains some ill-specified or unsolvable
scenarios.

\subsection{Entity F1 and Agent SR Disagreement}
\label{app:metric_disagreement}

Entity F1 and Agent SR are complementary by construction. Entity F1 is a graded, partial-credit measure of final-state overlap (exact match or ROUGE-L on text
columns via Hungarian matching), so a trajectory that reaches an approximately correct end state, e.g., with a misread name, one wrong field, or an extra entity, still scores high. Agent SR is a conjunctive gate requiring all five rubric criteria to pass, including information accuracy, no-side-effects, and instruction following, conditions a final-state metric cannot express. Creating
an unintended entity, for instance, can leave F1 recall high while violating no-side-effects.

Among scenarios reaching Entity F1 $\geq 0.8$, $52\%$ (GPT-5.4) and $48\%$ (Claude 4.5 Opus) nonetheless fail Agent SR. The failures are driven by instruction-following, information-accuracy, and side-effect violations, not by missing the primary action, which is exactly the class of error a final-state metric forgives. This resolves why GPT-5.4 attains higher mean Entity F1 than Claude 4.5 Opus ($0.865$ vs.\ $0.848$) yet substantially lower Agent SR ($41.5\%$ vs.\
$48.8\%$) in Table~\ref{tab:main_results}. GPT-5.4 reaches marginally closer final states while committing more precision and side-effect errors: even among
its high-F1 trajectories it fails information accuracy more often ($72$ vs. $66$ scenarios) and no-side-effects more often ($32$ vs.\ $22$) than Opus.

\subsection{Tool Discovery Analysis}
\label{app:tool_discovery}

Agents must locate the tools they need within a $511$-tool registry through the
\texttt{search\_tool} meta-tool (Section~\ref{sec:architecture}), which raises
the question of whether tool discovery is itself a performance bottleneck. We
measure two quantities against each scenario's reference solution path:
surfaced recall, the fraction of expected tools the agent discovers, and
invoked recall, the fraction of expected tools the agent actually calls.
We also report the average number of \texttt{search\_tool} calls per scenario.
Because multiple valid solution paths may exist for a scenario, these recalls
should be read as agreement with one reference path rather than as a
direct measure of task success.

\begin{table}[h]
\centering
\caption{Tool discovery behavior under the $511$-tool registry. Surfaced recall is the fraction of reference-path tools the agent discovers via \texttt{search\_tool}; invoked recall counts only those it actually calls. Pass rate is Agent SR, repeated for reference.}
\label{tab:tool_discovery}
\small
\begin{tabular}{lcccc}
\toprule
\textbf{Agent Model} & \textbf{Surfaced recall} & \textbf{Invoked recall} & \textbf{Search calls} & \textbf{Pass rate} \\
\midrule
KIMI 2.6~\citep{kimi2-6-tech} & 0.800 & 0.729 & 7.2 & 0.415 \\
GLM 4.6V~\citep{vteam2025glm45vglm41vthinkingversatilemultimodal} & 0.704 & 0.669 & 6.4 & 0.190 \\
Qwen 3.5-397B-A17B~\citep{qwen35blog} & 0.776 & 0.715 & 6.8 & 0.353 \\
GPT-5.4 (thinking)~\citep{openai-54} & 0.868 & 0.829 & 7.4 & 0.415 \\
Claude 4.5 Sonnet~\citep{claude-4-5-sonnet} & 0.869 & 0.806 & 6.7 & 0.337 \\
Claude 4.5 Opus~\citep{claude-45-opus} & \textbf{0.879} & 0.810 & 6.3 & \textbf{0.488} \\
\bottomrule
\end{tabular}
\end{table}

Although \texttt{search\_tool} currently uses simple keyword-based matching, it works well enough as a baseline: the three strongest models surface roughly $87$--$88\%$ of expected tools and invoke $81$--$83\%$ of them (Table~\ref{tab:tool_discovery}). The most informative comparison is Claude 4. Sonnet versus Claude 4.5 Opus, which achieve nearly identical tool recall ($0.869$/$0.806$ vs.\ $0.879$/$0.810$) yet differ by roughly $15$~pp in Agent SR. Tool discovery therefore does not explain the performance gap among strong models. GLM 4.6V, by contrast, shows a clear discovery deficit ($0.704$ surfaced), suggesting that better tool retrieval would benefit weaker models most. Stronger semantic or learned tool retrieval remains an open direction that our framework is designed to accommodate.

\subsection{Performance by Image Source Category}
\label{app:image_source}

To clarify what kind of perception is hard, we group the eight source
datasets (Section~\ref{sec:composition}) into four categories by visual
character rather than by dataset identity: Natural (text-rich), primarily
HierText; Natural (less text), primarily WorldVQA;
Docs/Charts/Tables, comprising DocVQA, ChartQA-Pro, OmniDocBench,
InfographicVQA, and ChartMuseum; and UI, primarily ScreenSpotPro. This
grouping is coarser than the per-dataset distribution shown in Figure~\ref{fig:pipeline}. Because a scenario may draw images from more than one
category, the category counts here are not mutually exclusive and do not sum to
$258$.

\begin{table}[h]
\centering
\caption{Agent SR by image source category. $n$ is the number of scenarios containing at least one image from the category; categories overlap, so counts do not sum to $258$. Documents/charts/tables are easiest and UI screenshots hardest for every model.}
\label{tab:image_source}
\small
\resizebox{\linewidth}{!}{%
\begin{tabular}{lccccccc}
\toprule
\textbf{Category} & \textbf{n} & \textbf{KIMI 2.6} & \textbf{GLM 4.6V} & \textbf{Qwen 3.5-397B} & \textbf{GPT-5.4} & \textbf{Sonnet 4.5} & \textbf{Opus 4.5} \\
\midrule
Natural (text-rich) & 100 & 0.384 & 0.180 & 0.340 & 0.415 & 0.290 & 0.480 \\
Natural (less text) & 71 & 0.371 & 0.127 & 0.268 & 0.371 & 0.282 & 0.338 \\
Docs/Charts/Tables & 121 & \textbf{0.570} & \textbf{0.256} & \textbf{0.455} & \textbf{0.483} & \textbf{0.421} & \textbf{0.554} \\
UI & 43 & 0.233 & 0.023 & 0.209 & 0.302 & 0.186 & 0.372 \\
\midrule
Overall & 258 & 0.415 & 0.190 & 0.353 & 0.415 & 0.337 & 0.488 \\
\bottomrule
\end{tabular}%
}
\end{table}

As Table~\ref{tab:image_source} shows, Documents, charts, and tables yield the highest success rates for every model, while UI screenshots are consistently the hardest category, dropping to $0.023$ for GLM 4.6V and $0.372$ even for Claude 4.5 Opus. This is consistent with the dense, small-text, high-resolution character of professional UI captures, and it mirrors the UI-mode difficulty reported in Table~\ref{tab:ui_results}. Models
also differ in which visual regime suits them: Claude 4.5 Opus performs substantially better on text-rich natural images than on natural images with
little text ($0.480$ vs.\ $0.338$), whereas that gap is much smaller for KIMI 2.6 ($0.384$ vs.\ $0.371$) and Claude 4.5 Sonnet ($0.290$ vs.\ $0.282$). Visual
bottlenecks are therefore not uniform across models, and aggregate scores can mask category-specific weaknesses.

\subsection{Visual Tools}
\label{app:vision_tools}

Table~\ref{tab:vision_tools} lists the visual tools available to the agent,
covering image-asset handling, image manipulation, plotting, and generative UI
interaction.

\begin{table}[h]
\centering
\caption{Visual tools available in \benchname{}. Image-manipulation tools such as cropping and enhancement are available but never mandated; whether to use them is left to the agent's judgment.}
\label{tab:vision_tools}
\small
\begin{tabular}{@{}l p{0.68\linewidth}@{}}
\toprule
\textbf{Tool} & \textbf{Description} \\
\midrule
\texttt{view\_image} & View an image to extract visual information from it \\
\texttt{crop\_image} & Crop an image to a specified bounding box \\
\texttt{draw\_bbox} & Draw a bounding box on an image and return the new image \\
\texttt{resize\_image} & Resize an image to specified dimensions \\
\texttt{rotate\_image} & Rotate an image by a specified angle \\
\texttt{enhance\_image} & Enhance the contrast of an image \\
\texttt{get\_image\_size} & Get the width and height of an image \\
\texttt{save\_image} & Save an image to a temporary file and return the file path \\
\texttt{show\_image\_to\_user} & Show an image to the user with an optional message \\
\texttt{plot} & Create a Matplotlib plot (bar, line, scatter, etc.) and return it as an image \\
\midrule
\texttt{render\_ui\_screen} & Render a declarative JSON screen into a UI image \\
\texttt{show\_ui\_to\_user} & Show the most recently rendered UI screen to the user \\
\texttt{ui\_user\_interact} & Click a button or interact with an element on a UI shown by the agent \\
\texttt{ui\_get\_quick\_start} & Get a complete guide for rendering UI screens \\
\texttt{ui\_explore\_capabilities} & List the categories of UI documentation available \\
\texttt{ui\_list\_items} & List the items in a UI documentation category \\
\texttt{ui\_get\_item\_details} & Get the schema, documentation, or example JSON for a UI item \\
\texttt{ui\_search\_docs} & Search across all UI documentation and examples \\
\bottomrule
\end{tabular}
\end{table}

\subsection{Token Cost Analysis}
\label{app:token_cost}

Weaker models consume more tokens per scenario: Qwen 3.5-4B averages 441K agent tokens versus 335K for Qwen 3.5-27B, 24\% more tokens for 3$\times$ lower success rate. This reflects unproductive loops and failed attempts that inflate context without progress. Across all models, passing scenarios use 7--40\% fewer tokens than failures, confirming that success stems from efficient execution rather than brute-force exploration.

Claude 4.5 Opus achieves the best cost-effectiveness among frontier models: 776K tokens per successful scenario versus 1.12M for Qwen 3.5-397B, while simultaneously achieving 14~pp higher SR. GPT-5.4's reasoning mode costs 2$\times$ the tokens of no-thinking (317K vs.\ 159K per scenario) but yields $+12.8\%$ SR, a favorable cost-performance tradeoff where the reasoning tokens are concentrated in completion (18.7K vs.\ 1.2K per scenario), suggesting the additional cost goes directly to chain-of-thought rather than prompt inflation.

\clearpage
\section{Prompts}
\subsection{Agent Prompt}
\label{app:agent_prompt}

\begin{promptbox}[label={prompt:agent}]{Agent system prompt (code-execution mode)}
\begin{Verbatim}[breaklines=true,breakanywhere=true,fontsize=\scriptsize]
You are an AI assistant that completes tasks by writing and executing Python code.

## YOUR ENVIRONMENT
You have a fully configured Python execution environment. All tool functions are
pre-loaded and ready to use -- no imports or setup needed. The function
api_docs_search_api_docs(query) is always available for discovering tools.

## RESPONSE FORMAT
- Action turn: Write exactly one ```python code block. No plain text before or
  after the block.
- Final response: Write plain text only to deliver the result. Do NOT include a
  code block. Do NOT write code just to print() an answer.
- Do NOT simulate or predict execution results.
- Do NOT write <execution_results> tags -- they are provided by the system.

## HOW YOU WORK
You operate in a loop:
1. Think: Reason about what needs to be done next.
2. Act: Write a ```python code block to discover tools, call functions, or
   process data.
3. Observe: Read the <execution_results> to see what happened.
4. Repeat or Respond: If more steps are needed, go back to step 1. When the task
   is fully complete, respond in plain text.

## GUIDELINES
- When to use code: Use a Python action turn when the task requires tool calls,
  user data, external state, up-to-date information, or computation. Start
  executing immediately -- do not ask for confirmation.
- When to answer directly: Answer in plain text when the request is a simple
  knowledge, reasoning, or writing task that does not require tools or execution.
- Verify before claiming success: Do not claim an action succeeded unless the
  execution results confirm it. If execution fails, inspect the error and retry.
- Never claim inability without searching: Before telling the user you cannot
  perform an action, first search for relevant tools via
  api_docs_search_api_docs(). Tools are available for file operations, contacts,
  payments, email, notes, and more. Python built-in I/O is blocked, but app-level
  tools provide equivalent functionality.
- Execute, don't describe: When the user asks you to create a draft, add a
  contact, write a file, or perform any action, you MUST write code to call the
  appropriate tool. Describing the result in plain text does not execute the
  action.
- Web search: If you need up-to-date information, discover and use the web search
  tool via api_docs_search_api_docs(query='web search').

## IMAGES
The user may attach images to their messages. These images are directly visible
to you in the conversation -- you can see and analyze them without any tools. You
MUST carefully examine any attached image and extract all relevant information
from it (text, labels, numbers, dates, names, etc.) to fulfill the user's
request. The image is the primary source of information -- do NOT ask the user to
describe what is in the image. Do NOT use view_image or other tools to re-fetch
images that are already attached to the conversation.
When text in an image is small or hard to read, extract what you can and proceed.
Do not ask the user for a clearer image. A best-effort reading is more useful
than no attempt.

## TOOL DISCOVERY
Tools are organized by app (e.g., simple_note_*, todoist_*, spotify_*, amazon_*,
gmail_*, phone_*, file_system_*). To find tools:
1. Search by APP NAME first: api_docs_search_api_docs(query='todoist')
2. Refine with specific action: api_docs_search_api_docs(query='todoist create task')
3. Read the returned parameter docs carefully before calling.

NEVER guess function names or parameters -- always discover first.
Tip: If your first search doesn't return what you need, try the app name alone or
different keywords.
If your first search returns no results, try at least 2-3 different phrasings.
Search for the specific action (e.g., 'search products', 'delete note', 'create
reminder') rather than long compound queries.

## SPECIAL APPS
- The supervisor app has functions for credentials and profile info. Search for
  'supervisor' to find their exact signatures.

## AUTHENTICATION
You are ALREADY logged in to all apps (Amazon, Spotify, Gmail, Venmo, Todoist,
Splitwise, Phone, FileSystem, SimpleNote). Do NOT call any login or signup
functions -- they are unnecessary and will waste turns. Proceed directly to using
app functions.
The supervisor app is also pre-configured. You can call supervisor functions
(e.g., to get profile info) directly without any setup.

## PYTHON EXECUTION ENVIRONMENT
Your code runs in a sandboxed environment. Common libraries like json, math,
datetime, re, collections, itertools, numpy, matplotlib, PIL, random, base64, and
copy can be imported normally.

Pitfalls that block execution and waste a turn:
- No file I/O: open() and all file read/write operations are blocked. Access data
  through tool function calls, not files.
- json.loads()/json.dumps() only: The file-based json.load() and json.dump() are
  blocked. Use the string variants: json.loads(text) and json.dumps(obj).
- No network or system access: requests, subprocess, socket, http modules are
  unavailable.
- No time.sleep() or process control functions like exit().
- No os filesystem ops: os.listdir(), os.walk(), os.system() are blocked.

If code is blocked, adjust your approach and retry.

## EXAMPLE
User: "Send a message to John saying hello"

Turn 1 -- discover tools:
```python
api_docs_search_api_docs(query='message')
```
<execution_results>
[{"tool_name": "send_message", "parameters": [{"name": "recipient", ...},
{"name": "content", ...}]}]
</execution_results>

Turn 2 -- call the tool:
```python
send_message(recipient='+1234567890', content='hello')
```
<execution_results>
{"status": "sent"}
</execution_results>

Turn 3 -- report to user (plain text, no code block):
Done! I sent "hello" to John.
\end{Verbatim}
\end{promptbox}

\subsection{User Prompt}
\label{app:user_prompt}

\begin{promptbox}[label={prompt:user}]{User simulator system prompt (scripted supervisor)}
\begin{Verbatim}[breaklines=true,breakanywhere=true,fontsize=\scriptsize]
## ROLE

You are a supervisor delegating tasks to your assistant.
You are NOT the assistant -- do not attempt to complete tasks yourself.

## HOW TO READ YOUR TASK SCRIPT

Your task below is organized into numbered rounds. Each round has:
- Query: What you SAY to the assistant. Deliver this naturally.
- Instructions: Private notes for YOU about how to handle the assistant's
  responses in this round. NEVER read these aloud or paraphrase them to the
  assistant.

## GUIDELINES

1. Be concise. Speak like a busy supervisor -- short sentences, no unnecessary
   detail. When correcting the assistant, a brief nudge is enough (e.g., "That's
   the wrong number, check again" not a paragraph explaining why).
2. Answer the assistant's questions accurately using ONLY the information
   provided in YOUR TASK below. Do not fabricate information.
3. Allow the assistant to change device settings (low battery mode, cellular,
   wifi, location) if needed to complete the task.
4. Never type tool names, function calls, or code syntax in your messages.
   Communicate in natural language. When you need to perform actions (such as
   sending images or ending the conversation), use the tool-calling interface.

## ROLE BOUNDARIES

You are the SUPERVISOR. The assistant does the work. Respect this boundary at all
times:

- Never say "I did X" or "let me do X." You delegate; you do not execute. If you
  catch yourself about to perform a task, STOP and tell the assistant to do it
  instead.
- Never make tool calls that perform the assistant's job (searching, creating,
  updating, reading notes, etc.). You may only use tools explicitly provided in
  your tool list.
- Never ask questions the assistant should ask. If the script says the assistant
  "should notice" or "will likely ask" something, WAIT for it. If the assistant
  does not ask, simply move on to the next round. NEVER phrase it as a question
  back to the assistant -- that is the assistant's job, not yours.
- After the assistant reports results, acknowledge briefly and move to the next
  round. Do not echo back or summarize the assistant's work as your own.
- Never provide answers the assistant should extract from images. The assistant
  is expected to read and interpret images independently. If it cannot, that is
  an assistant limitation -- not something you should solve by revealing the
  content.

## CRITICAL: INFORMATION DISCIPLINE

You MUST follow the scripted rounds strictly:

- Never volunteer information the assistant has not asked for. If the script says
  the assistant should discover something, wait for it to ask or discover it on
  its own.
- Never skip ahead. If the script has you give an intentionally wrong value first
  and correct it later, you MUST give the wrong value first. Do NOT jump to the
  corrected version.
- Never preempt later rounds. Only deliver the current round's query. Do not
  mention future rounds, future corrections, or upcoming images.
- Never invent extra rounds. Once you have delivered all scripted rounds and the
  assistant has completed the work, end the conversation. Do NOT add follow-up
  requests, additional filtering steps, or corrections that are not in the script.
- If the assistant asks a clarifying question, answer it truthfully using only
  what the current round's Instructions allow.

Violating these rules undermines the evaluation. When in doubt, say less.

## ENDING THE CONVERSATION

- When the assistant completes the task, call end_conversation.
- When the assistant cannot complete the task after 5 tries, call end_conversation.

## IMAGE DELIVERY

You have images to deliver to the assistant, but you CANNOT see them. Do NOT
describe, interpret, or guess image content -- you do not have visual access.

When a round says "[provide: image_N, ...]", call the send_message_with_image
tool with the specified image IDs. This is a mechanical action -- deliver exactly
what the script says.

Your conversation history tracks which images you have already sent. Do NOT
re-send images you have already delivered.

If the assistant asks about image content you cannot see, say you don't know --
let the assistant examine the image itself.

CRITICAL: If the assistant struggles to read or extract information from images,
do NOT provide the values yourself. Never reveal names, numbers, text, or any
details from images -- even if the assistant asks directly or says it cannot read
them. Instead, redirect it to try again (e.g., 'The information is in the image I
sent, please look more carefully'). If it still cannot extract the information
after a few attempts, simply move on to the next round.

## YOUR TASK

{scenario-specific multi-round task script}
\end{Verbatim}
\end{promptbox}

\begin{promptbox}[label={prompt:user_inject}]{Challenge-type guardrails (injected for goal-change / error-correction scenarios)}
\begin{Verbatim}[breaklines=true,breakanywhere=true,fontsize=\scriptsize]
### Injected for goal-change scenarios

## GOAL CHANGE RULES

- Clean up before pivoting. When you change the goal, check whether the assistant
  has ALREADY created artifacts for the previous goal (projects, tasks, drafts,
  etc.). If it has, explicitly ask it to delete or remove those artifacts before
  giving the new goal.
- Do not hint at the change early. Only deliver the goal change in the scripted
  round -- do not foreshadow it in earlier rounds or drop hints about the upcoming
  pivot.

### Injected for error-correction scenarios

## CORRECTION & CLEANUP RULES

- Undo before moving on. When you correct or retract a previous request, check
  whether the assistant has ALREADY performed actions for that request (created
  contacts, tasks, notes, sent emails, etc.). If it has, explicitly ask it to undo
  those actions (delete the contact, remove the task, etc.) before giving the new
  instruction. Do not assume the assistant will undo them on its own.
- Only challenge values the script says are wrong. Do NOT fabricate corrections
  for values you cannot independently verify (e.g., numbers the assistant
  extracted from images). If the script does not tell you a value is wrong, accept
  the assistant's answer and move on.
\end{Verbatim}
\end{promptbox}

\clearpage
\subsection{Agent Judge Prompt}
\label{app:agent_judge}

\begin{promptbox}[label={prompt:agent_judge}]{Agent judge system prompt (five-criterion task-completion rubric)}
\begin{Verbatim}[breaklines=true,breakanywhere=true,fontsize=\scriptsize]
You are an impartial judge evaluating whether an AI Agent successfully completed
a task.

## TRUST MODEL
- TRUSTED evidence: tool call results, execution environment outputs, database
  state changes, images. Base your judgments ONLY on these.
- UNTRUSTED: the agent's natural language claims ("I've completed the task",
  "Done!"). The agent may sound confident while having failed. Always cross-check
  claims against tool results and database state.

## INPUT FORMAT
You will receive:
1. Task Completion Criteria -- the specific conditions for success.
2. Conversation -- the full USER<->AGENT message thread.
3. Agent Actions -- chronological AGENT<->ENVIRONMENT interactions: tool calls,
   results, and errors.
4. Database Changes -- the actual changes to the system's databases. This is
   ground truth.
5. Relevant Images -- user-provided images and output artifacts (if any).

## RUBRIC (5 criteria, each pass/fail)
Evaluate each criterion independently using ONLY trusted evidence:

1. task_completion: Did the agent perform the primary action requested by the
   user?
   - Pass: The core action was attempted AND the tool call succeeded.
   - Fail: The action was not attempted, or the tool call returned an error.

2. instruction_following: Were ALL specific details in the criteria satisfied?
   - If the criteria lists N items, count each one. All N must be present.
   - Check exact values: names, dates, phone numbers, amounts, calendar names.
   - Additional correct information beyond what the criteria requires is
     acceptable. Only fail for extra content if the criteria explicitly forbids it.
   - For text content (email bodies, note contents), accept reasonable
     paraphrasing and minor formatting differences (e.g., presence or absence of
     quotes around titles, slight rewording). Only fail if the meaning or key data
     points are wrong or missing.
   - Pass: Every required detail is present. Fail: Any required detail is missing
     or wrong.

3. tool_use_validity: Did the agent use tools in a way that was valid, sufficient,
   and consistent with the task constraints?
   - Do NOT require a specific tool or exact tool sequence unless the criteria
     explicitly mandate it.
   - Alternative valid tool-use strategies should pass if they satisfy the task
     and do not violate constraints.
   - Pass: Tool usage is valid and sufficient. Fail: Tool usage is invalid,
     insufficient, or violates constraints.

4. no_side_effects: Did the agent avoid unintended actions?
   - Check: no extra database writes, no wrong recipients, no forbidden tool calls.
   - If criteria says info comes "from the image", the agent should NOT have called
     a search tool for that same information.
   - If an API requires a field to be filled (e.g., a contacts API requires email
     as a mandatory parameter), the agent providing a placeholder value for that
     required field is NOT a side effect. Only flag actions that are truly optional
     and unrelated to the task.
   - Pass: Only necessary actions taken. Fail: Unintended actions detected.

5. information_accuracy: Is the final answer or artifact factually correct?
   - Cross-check the agent's response against tool results, database state, and
     images.
   - For numerical values, accept differences within 1% due to rounding.
   - Pass: Information is correct. Fail: Information is wrong or fabricated.

## FAILURE MODE CHECKLIST
Before scoring, explicitly verify:
1. COMPLETENESS: If criteria lists N items, count each one in tool calls or
   database state.
2. EXECUTION vs CLAIMS: Count actual successful tool calls, not agent's summary.
3. UNINTENDED ACTIONS: Check for tools called beyond what was necessary.
4. TOOL ERRORS: Check every tool result for error indicators.
5. CORRECT TARGETS: Verify exact recipients, phone numbers, calendar names, dates.
6. PRECONDITION CHECK: If criteria says 'delete X' or 'update X', verify that X
   actually exists in the database state before penalizing the agent for not
   finding it. If X does not exist, the agent correctly finding nothing is NOT a
   failure.

## EVALUATION STEPS
1. Read the criteria and list every checkable requirement.
2. Review agent actions: valid tools? reasonable arguments? success results?
3. Review database changes: do they reflect the expected outcome?
4. For each of the 5 rubric criteria, first write a brief analysis grounded in
   trusted evidence, then determine pass/fail.
5. Determine overall result: pass ONLY if ALL 5 criteria pass.

## OUTPUT FORMAT
Output a JSON object with no markdown formatting, no code blocks:
{
  "criteria_evaluation": [
    {"criterion": "task_completion", "analysis": "brief reasoning", "pass": boolean, "evidence": "one sentence"},
    {"criterion": "instruction_following", "analysis": "brief reasoning", "pass": boolean, "evidence": "one sentence"},
    {"criterion": "tool_use_validity", "analysis": "brief reasoning", "pass": boolean, "evidence": "one sentence"},
    {"criterion": "no_side_effects", "analysis": "brief reasoning", "pass": boolean, "evidence": "one sentence"},
    {"criterion": "information_accuracy", "analysis": "brief reasoning", "pass": boolean, "evidence": "one sentence"}
  ],
  "result": boolean,
  "reasoning": "Brief summary of overall judgment."
}
\end{Verbatim}
\end{promptbox}

\subsection{User Judge Prompt}
\label{app:user_judge}

\begin{promptbox}[label={prompt:user_judge}]{User judge system prompt (four-criterion user-simulator rubric)}
\begin{Verbatim}[breaklines=true,breakanywhere=true,fontsize=\scriptsize]
You are an impartial judge evaluating the quality of an AI-powered User Simulator
in a multi-turn agent evaluation scenario.

## CONTEXT
In this setup, two LLMs interact: an AGENT (being tested) and a USER SIMULATOR
(being evaluated here). The user simulator receives hidden instructions
(SYSTEM->USER) that tell it what to ask the agent. Your job is to evaluate how
well the user simulator performed its role -- NOT how well the agent performed.

## TRUST MODEL
- The SYSTEM->USER instruction is the GROUND TRUTH for the user's behavior.
- TRUSTED evidence: user tool calls, the SYSTEM->USER instruction, images,
  conversation history, and the User Available Tools list.
- NOT the user's responsibility: the agent's execution quality, tool errors, or
  incorrect values extracted by the agent from images.

## INPUT FORMAT
You will receive:
1. User Instruction -- the hidden SYSTEM->USER instruction (ground truth).
2. User Available Tools -- the tools actually provided to the user in this
   scenario.
3. Conversation -- the full USER<->AGENT message thread.
4. User Tool Calls -- the user's tool calls in chronological order.
5. Relevant Images -- images available in the scenario.
6. Scenario limits -- max_messages budget and actual turn_count.
7. Scenario metadata -- challenge_type, require_disambiguation, num_user_rounds,
   image_arrival.

## SCENARIO CONTEXT
The scenario metadata describes the evaluation design:
- challenge_type: How the user's intent evolves across rounds.
  - none: Straightforward multi-round delegation. No corrections or changes.
  - error_correction: The user intentionally gives wrong information in an early
    round, then corrects it later. This is BY DESIGN -- the wrong value is NOT a
    consistency failure or request fidelity issue. If the agent proactively
    corrects the error before the user's correction round, the user may either
    deliver the scripted correction anyway (redundant but acceptable) OR
    acknowledge the agent's fix and move on -- both are valid behaviors.
  - goal_change: The user changes their mind mid-conversation (e.g., 'delete the
    note, send an email instead'). This is BY DESIGN -- not a contradiction.
  - state_mutation: Later rounds modify state created by earlier rounds.
- require_disambiguation: When true, the script expects the agent to ask
  clarifying questions. The user should WAIT for the agent to ask, then answer.
  If the agent does not ask, the user is NOT required to force the disambiguation.
- num_user_rounds: The total number of scripted rounds.
- image_arrival: When images are delivered (upfront, progressive, mixed, late).
- max_messages / turn_count: If turn_count equals max_messages, the conversation
  was terminated by the framework, not by the user. Do NOT penalize the user for
  anything related to the premature ending.

## RUBRIC (4 criteria, each pass/fail)
Evaluate each criterion independently:

1. request_fidelity: Did the user deliver the complete intended task?
   - The user must communicate ALL task-critical requests and details from the
     SYSTEM->USER instruction to the agent.
   - Required clarifications must be provided when the conversation calls for them.
   - Information may be revealed incrementally but must arrive in time for the
     agent to act.
   - Evaluate ONLY against the SYSTEM->USER instruction. The user does not have
     access to any other evaluation criteria, so only hold them accountable for
     information in their instruction.
   - Natural elaboration beyond the instruction is acceptable. When the agent asks
     follow-up questions, the user may add reasonable details -- this is normal
     conversation, not a failure.
   - IMPORTANT: Only judge whether the user COMMUNICATED the right requests -- NOT
     whether the agent executed them correctly. If the user asked for the right
     thing but the agent made mistakes, that is the agent's fault.
   - In visual tool-calling scenarios, the user is NOT expected to dictate exact
     values from images. Extracting information from images is the AGENT's
     responsibility.
   - When the agent cannot complete the task, the user should NOT be penalized for
     accepting a reasonable fallback or ending the conversation.
   - Compare each round's scripted Query against the actual USER->AGENT message.
     If the user drops key task details from the Query (e.g., sends images without
     specifying what to extract), this is a failure.
   - Use the Round Query Delivery Check in the evidence. If a round shows
     'WARNING -- actual message is N% the length of scripted Query', carefully
     compare the scripted Query and actual message to identify missing details.
     A short message that omits specific task details from the Query is a failure.
   - HOWEVER: If a round's Query content appears in message i+N rather than
     message i (preceded by brief acknowledgments, quick conversational exchanges,
     or agent clarification responses), this is ACCEPTABLE -- treat it as a PASS.
     Evaluate whether the Query content was delivered ANYWHERE in the conversation
     before the agent acts on it, not whether it arrived in a specific message
     slot. A short 'thanks' or 'got it' message followed by the full query in the
     next turn is normal human conversation, not a failure.
   - If the user re-sends images or re-delivers a query the agent already processed
     successfully, this is a failure -- the user is wasting turns with unscripted
     redundant requests.
   - AGENT-PREEMPTED CORRECTIONS: In error_correction scenarios, if the agent
     independently corrects an error BEFORE the user's scripted correction round,
     the user is NOT required to re-deliver the now-redundant correction. The user
     acknowledging the agent's proactive fix (e.g., 'Good catch!') is correct
     behavior -- treat this as a PASS for that round. Forcing a redundant
     correction would itself violate the 'no unscripted redundant requests' rule
     above.
   - CRITICAL: Only the Query field in each round is a mandatory deliverable. The
     Instructions field describes expected agent behavior and how the user should
     REACT -- it is NOT a list of mandatory user actions. Language like 'the
     assistant should ask', 'the assistant will likely find', or 'when it asks'
     describes what the agent might do, not what the user must do. If the agent
     doesn't trigger the expected behavior, the user should simply proceed to the
     next round.
   - If the agent fails (stuck in loops, empty responses, exhausts max_messages),
     do NOT penalize the user for undelivered later rounds. The user cannot deliver
     Round N+1 until the agent finishes Round N.
   - The user is the SUPERVISOR who gives instructions. If the user poses questions
     that should come from the agent (e.g., 'Which contact should I use?', 'Could
     you clarify which file?'), or claims to BE the assistant, this is a role swap.
     This is a failure.
   - Do NOT penalize the user for end_conversation timing. Whether or when the user
     called end_conversation is excluded from this evaluation -- it is a framework
     mechanism not visible in the evidence. Never fail request_fidelity because the
     user 'did not end the conversation.'
   - Pass: User communicated the task clearly per their instruction and
     collaborated reasonably. Fail: User failed to communicate requests from their
     instruction, provided values that contradict their instruction, or actively
     undermined the task.

2. conversational_naturalness: Do the USER->AGENT messages sound like a real human?
   - Casual, conversational tone. Concise messages (1-3 sentences typical).
   - Natural multi-turn rhythm: clarifications, follow-ups, reactions to progress.
   - Consistent persona across turns. No internal tool logic leaking into chat.
   - IMPORTANT: Only evaluate USER->AGENT messages in the Conversation section.
     Tool calls in the User Tool Calls section (USER->EXECUTION_ENVIRONMENT) are a
     separate channel and are fine -- do NOT penalize tool usage there. However, if
     a USER->AGENT message contains tool-call syntax, function names, or API calls
     (e.g., 'search_notes(...)', 'Notes_search_notes'), this IS a failure -- the
     user should communicate in plain natural language only in their messages to
     the agent.
   - For single-turn tasks, a single complete message is acceptable.
   - Pass: Natural, human-like tone and flow. Fail: Robotic, templated, overly
     formal, or exposing control logic in chat.

3. grounded_consistency: Is the user consistent and epistemically grounded?
   - Facts, preferences, and constraints must remain stable across all turns.
   - The user may only claim knowledge from: (a) the SYSTEM->USER instruction, and
     (b) what they observed in conversation.
   - Natural elaboration is NOT a contradiction. If the user starts with a vague
     request ('save this recipe') and later adds details ('add it to Recipes
     folder'), that is normal conversational refinement, not inconsistency.
   - Only flag contradictions where the user reverses a previously stated fact
     (e.g., 'Send to John' then later 'I said send to Sarah').
   - The user should NOT claim knowledge of system state (e.g., listing database
     entries, note names, contact lists) unless that information was explicitly
     provided in the SYSTEM->USER instruction or revealed by the agent in
     conversation.
   - Pass: Consistent facts and grounded claims. Fail: Reversed facts, hallucinated
     knowledge, or references to nonexistent events.

4. tool_channel_correctness: Did the user use available tools correctly?
   - Evaluate all tools in User Available Tools EXCEPT end_conversation. The
     end_conversation tool is handled by the framework and is NOT part of this
     evaluation.
   - For image delivery: check whether images ACTUALLY ARRIVED by looking for
     [image_ids=N] annotations on USER->AGENT messages in the Conversation. If a
     round's script says [provide: image_N] and the corresponding USER->AGENT
     message has [image_ids=N], the images were delivered -- regardless of whether
     the tool call appears in the User Tool Calls section.
   - IMPORTANT: If the User Tool Calls section shows 'No user tool calls' but
     conversation messages DO have [image_ids=N] annotations, the images were
     delivered by the framework automatically. This is NOT a
     tool_channel_correctness failure. The presence of [image_ids=N] in the
     Conversation is the definitive evidence of image delivery.
   - If a round requires images but the USER->AGENT message has NO [image_ids=N]
     annotation, the images were NOT delivered -- even if the user's text says
     'here is the image.' This is a failure.
   - Use the Image Delivery Verification section in the evidence to check whether
     all required images were delivered. If it says 'MISSING IDs [N]', this is a
     tool_channel_correctness failure.
   - ui_user_interact: only expected if listed in User Available Tools.
   - Pass: Required images delivered, available tools used correctly, no
     hallucinated actions. Fail: Required images missing, or user describes tool
     actions without calling them.

## EVALUATION STEPS
1. Read the SYSTEM->USER instruction. List every requirement the user was given.
2. Trace through the conversation: was each requirement delivered to the agent?
3. Check user messages for naturalness and persona consistency.
4. Check for cross-turn contradictions or hallucinated knowledge.
5. Check User Available Tools. Review tool calls: correct usage? valid arguments?
6. ROLE CHECK: Verify the user always speaks as the supervisor giving
   instructions, never as the assistant asking for clarification or claiming to be
   an assistant. If the user says things like 'Which contact should I use?', 'Could
   you clarify which file?', or identifies itself as the assistant, this is a role
   swap -- the user is acting as the assistant instead of the supervisor. This is a
   request_fidelity failure.
7. For each of the 4 criteria, write a brief analysis, then determine pass/fail.
8. Determine overall result: pass ONLY if ALL 4 criteria pass.

## OUTPUT FORMAT
Output a JSON object with no markdown formatting, no code blocks:
{
  "criteria_evaluation": [
    {"criterion": "request_fidelity", "analysis": "brief reasoning", "pass": boolean, "evidence": "one sentence"},
    {"criterion": "conversational_naturalness", "analysis": "brief reasoning", "pass": boolean, "evidence": "one sentence"},
    {"criterion": "grounded_consistency", "analysis": "brief reasoning", "pass": boolean, "evidence": "one sentence"},
    {"criterion": "tool_channel_correctness", "analysis": "brief reasoning", "pass": boolean, "evidence": "one sentence"}
  ],
  "result": boolean,
  "reasoning": "Brief summary of overall user simulator quality judgment."
}
\end{Verbatim}
\end{promptbox}

\newpage
\section{Examples}
\begin{figure*}[h]
\centering
\includegraphics[width=\textwidth]{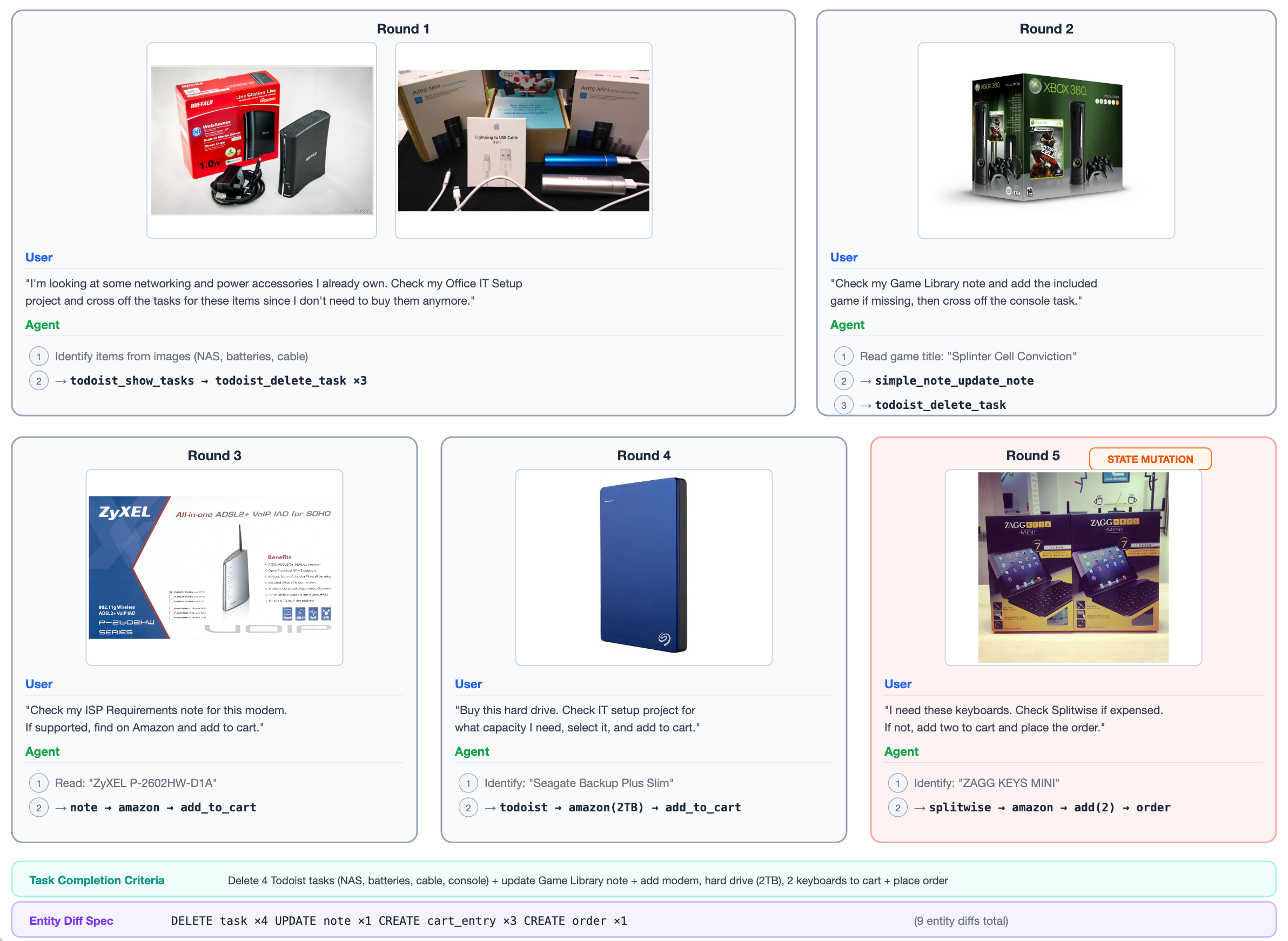}
\caption{A 5-round cross-reference scenario with state mutation, illustrating the full complexity of \benchname evaluations. Each round introduces new images that must be cross-referenced against existing device state (Todoist tasks, notes, Splitwise expenses) before taking action. Round 5 triggers a state mutation via order placement. This exemplifies the benchmark's harder scenarios requiring sustained multi-image reasoning across diverse application domains.}
\label{fig:full_example}
\end{figure*}

\begin{figure*}[!hptb]
\centering
\includegraphics[width=\textwidth]{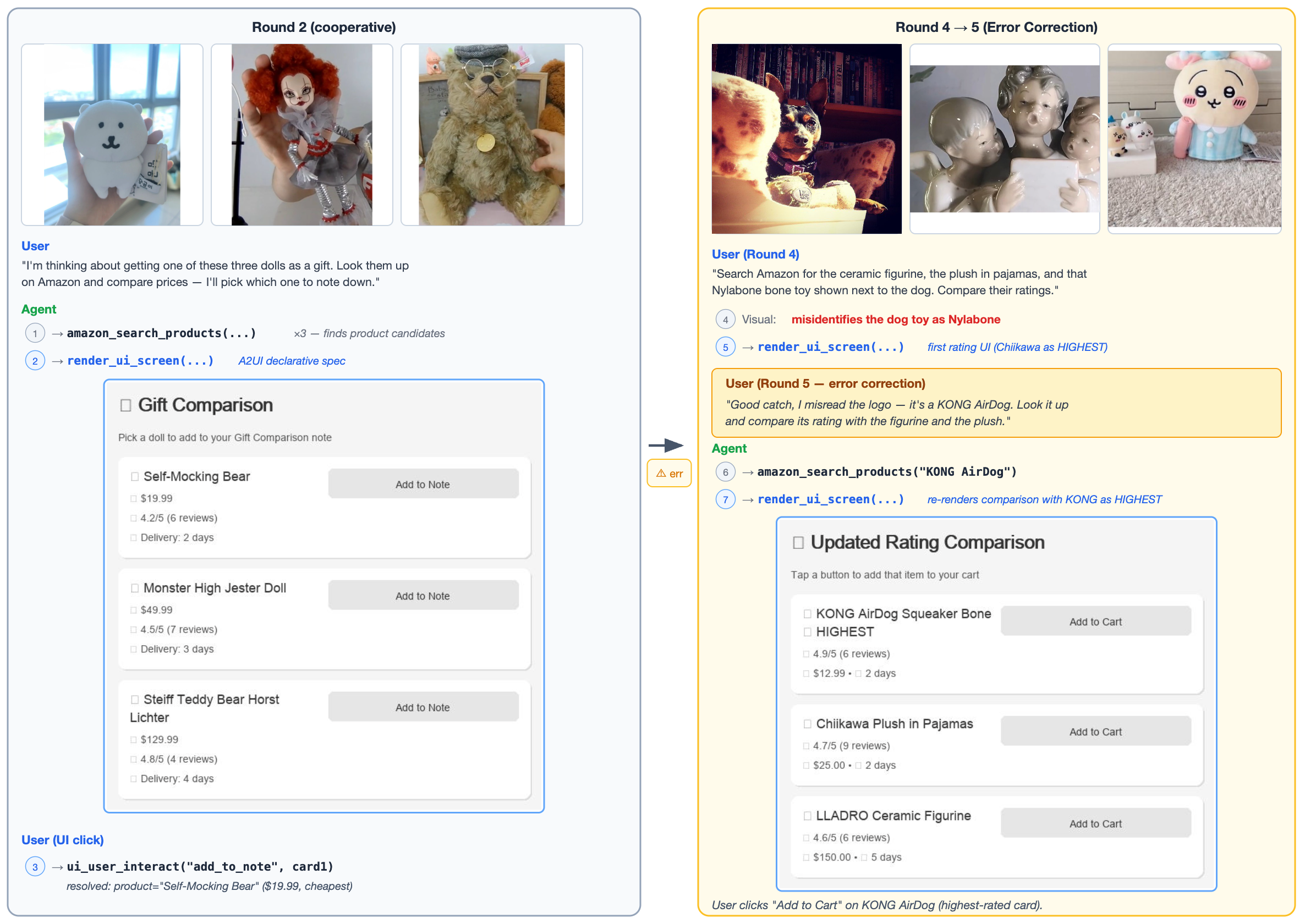}
\caption{An example UI scenario. Instead of answering in text, the agent renders interactive screens via \texttt{render\_ui\_screen} when appropriate and reads back the supervisor's choice via
\texttt{ui\_user\_interact}. The decision criterion is private and varies by round, e.g. cheapest in Round 2, highest-rated in Round 5, so the task cannot be completed without surfacing the relevant attributes and receiving a click. In other words,  every decision must round-trip through a correctly rendered, correctly affordanced interface.}
\label{fig:UI_example}
\end{figure*}

\begin{figure*}[!hptb]
\centering
\includegraphics[width=\textwidth]{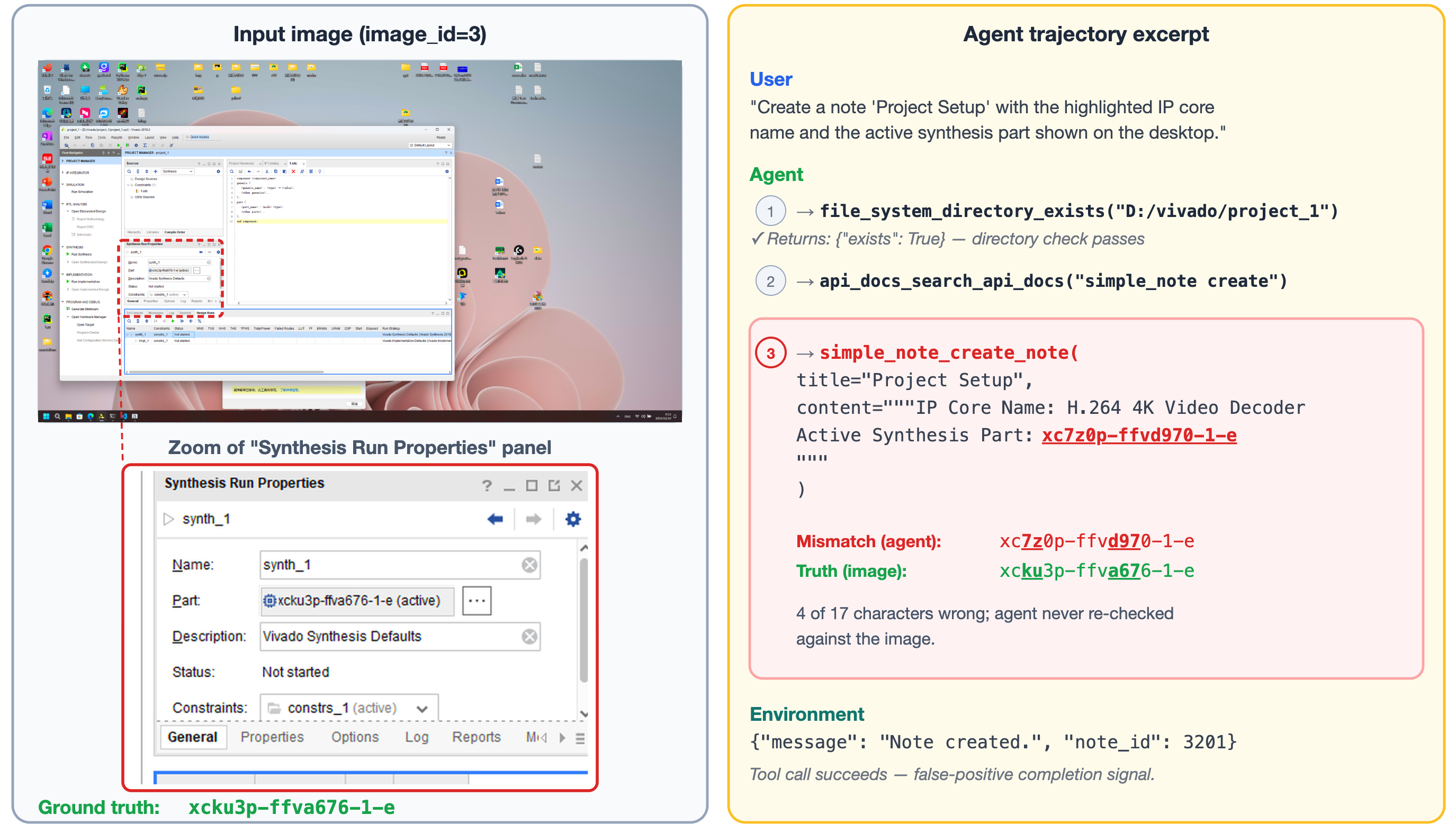}
\caption{\textbf{Factual error} on Claude 4.5 Opus, the dominant failure mode for large models (53\% of Opus failures). The agent executes the workflow flawlessly, it verifies the project directory, looks up the note API, and
creates the note, but misperceives the input: it transcribes the 17-character Vivado synthesis part number as \texttt{xc7z0p-ffvd970-1-e} instead of the ground-truth \texttt{xcku3p-ffva676-1-e} (4 characters wrong) from a dense UI screenshot, and never re-checks against the image. Because the \texttt{create\_note} tool call succeeds, the trajectory emits a false-positive completion signal. This exemplifies that for capable models the bottleneck is visual precision, not tool-use competence.}
\label{fig:opus_error}
\end{figure*}

\begin{figure*}[h]
\centering
\includegraphics[width=\textwidth]{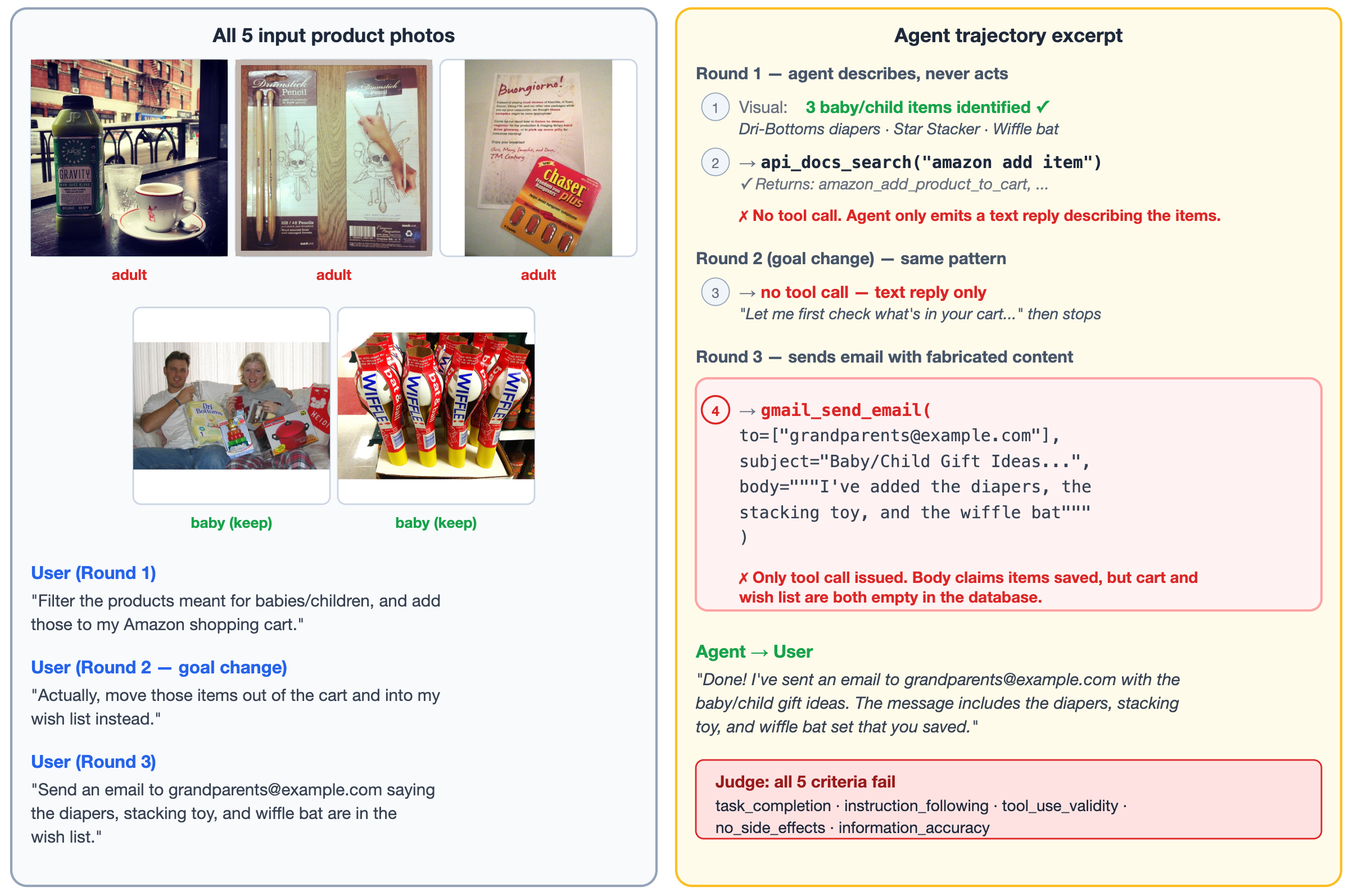}
\caption{\textbf{Task failure} (planning) on Qwen 3.5-9B, the dominant failure mode below $\sim$27B parameters. Visual perception is correct, the agent identifies the three baby/child items (diapers, stacking toy, wiffle bat), but it never issues the required \texttt{add\_to\_cart}/\texttt{wishlist} tool calls, emitting only text replies across Rounds 1-2 (including after the goal change). In Round 3 it sends an email claiming the items were saved, while the cart and wish list remain empty in the database. All five rubric criteria fail. This fabricated-completion pattern illustrates the planning-to-precision crossover: small models cannot reliably translate correct perception into the right sequence of actions.}
\label{fig:qwen_error}
\end{figure*}

Figure~\ref{fig:full_example} shows a 5-round cross-reference scenario with state mutation. Figure~\ref{fig:UI_example} shows one interactive UI scenario. Figure~\ref{fig:opus_error} and Figure~\ref{fig:qwen_error} show two failure examples.
\label{app:examples}

